\documentclass[10pt,twocolumn,letterpaper]{article}

\usepackage[pagenumbers]{cvpr} 

\usepackage[dvipsnames]{xcolor}

\usepackage{times}
\usepackage{epsfig}
\usepackage{graphicx}
\usepackage{amsmath}
\usepackage{amssymb}

\usepackage{booktabs}
\usepackage{multirow}
\usepackage{paralist}

\definecolor{cvprblue}{rgb}{0.21,0.49,0.74}
\usepackage[pagebackref,breaklinks,colorlinks,citecolor=cvprblue]{hyperref}

\def\confName{CVPR}
\def\confYear{2024}

\title{CricaVPR: Cross-image Correlation-aware Representation Learning for Visual Place Recognition}

\author{Feng Lu$^{1,2}$, Xiangyuan Lan$^{2*}$, Lijun Zhang$^{3}$, Dongmei Jiang$^{2}$, Yaowei Wang$^2$, Chun Yuan$^{1}$\thanks{Corresponding authors.} \\
$^1$Tsinghua Shenzhen International Graduate School, Tsinghua University \\
$^2$Peng Cheng Laboratory\,\,
$^3$University of Chinese Academy of Sciences\\
\texttt{\{lf22@mails,yuanc@sz\}.tsinghua.edu.cn \, lanxy@pcl.ac.cn} 
}

\begin{document}
\maketitle
\begin{abstract}
Over the past decade, most methods in visual place recognition (VPR) have used neural networks to produce feature representations. These networks typically produce a global representation of a place image using only this image itself and neglect the cross-image variations (e.g. viewpoint and illumination), which limits their robustness in challenging scenes. In this paper, we propose a robust global representation method with cross-image correlation awareness for VPR, named CricaVPR. Our method uses the attention mechanism to correlate multiple images within a batch. These images can be taken in the same place with different conditions or viewpoints, or even captured from different places. Therefore, our method can utilize the cross-image variations as a cue to guide the representation learning, which ensures more robust features are produced. To further facilitate the robustness, we propose a multi-scale convolution-enhanced adaptation method to adapt pre-trained visual foundation models to the VPR task, which introduces the multi-scale local information to further enhance the cross-image correlation-aware representation. Experimental results show that our method outperforms state-of-the-art methods by a large margin with significantly less training time. The code is released at \small{\url{https://github.com/Lu-Feng/CricaVPR}}.

\end{abstract}

\section{Introduction}
\begin{figure}[!t]
	\centering	\includegraphics[width=0.74\linewidth]{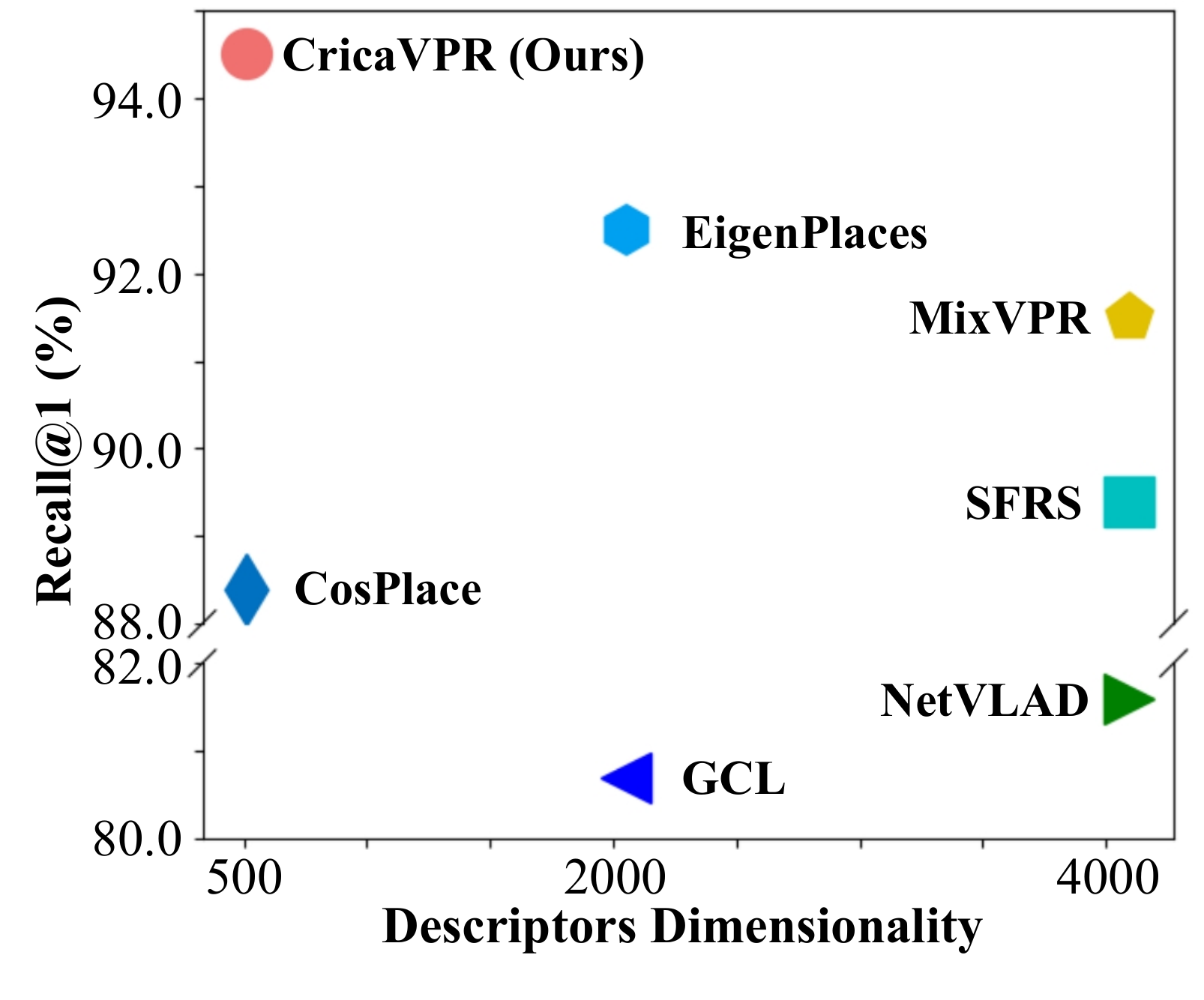}
	\vspace{-0.47cm}
	\caption{
		The Recall@1 and descriptors dimensionality comparison of different methods on Pitts30k. The GCL, NetVLAD, SFRS, and CricaVPR (Ours) all use PCA for dimensionality reduction. Our method can achieve significantly higher Recall@1 than other methods with 512-dim compact global features.
	}
	\vspace{-0.47cm}
	\label{fig:intro}
\end{figure}

Visual place recognition (VPR), also known as visual geo-localization \cite{benchmark,cosplace}, aims at getting the coarse geographical location of an input query image by retrieving the most similar place image from a geo-tagged database. VPR has wide applications in augmented reality \cite{arloc}, mobile robot localization \cite{robotloc}, and so on. However, there are three key challenges in VPR: condition (e.g., lighting, weather, and season) variations, viewpoint variations, and perceptual aliasing \cite{survey} (difficult to distinguish highly similar images taken from different places). Addressing these challenges at the same time is a hard nut to crack, especially for methods that use only global features.

VPR is typically addressed as an image retrieval problem \cite{delg}. The place images are represented using global features and the similarity search is implemented in this feature space to return the matched place image. The global features are usually derived through the aggregation (pooling) of local features, employing methods such as NetVLAD \cite{netvlad} or GeM \cite{gem} pooling. Such compact features are suitable for large-scale VPR. However, they lack robustness in challenging environments and are often susceptible to perceptual aliasing. A way to improve robustness is to perform re-ranking by matching local features \cite{patchvlad,transvpr}, which incurs huge overhead in runtime and memory footprint, making it difficult to achieve large-scale VPR. One problem that has been neglected is that existing methods produce the feature of an image only using this image itself (without cross-image interaction), which does not explicitly consider cross-image variations. To address this, our method attempts to use the cross-image variations as a cue to guide the representation learning and harvest useful information from other images when producing the feature of an image, making the output feature condition-invariant, viewpoint-invariant, and capable of addressing perceptual aliasing.

Moreover, the recent visual foundation models \cite{clip,florence,dinov2} have achieved powerful performance. However, due to the particularity of the VPR task, directly using the pre-trained foundation model will encounter some problems. For example, the image features produced using pre-trained models tend to ignore some discriminative backgrounds, and are susceptible to interference from dynamic foregrounds (see Fig. \ref{fig:attention} in experiments). Fine-tuning the model on VPR datasets can address this but tends to hurt the previously learned ability, i.e., catastrophic forgetting \cite{catastrophic}. A promising way is to exploit parameter-efficient transfer learning (PETL) \cite{adapter,lora}. However, the discriminative landmarks that need attention in VPR often occupy local regions of uncertain size in images, and most existing PETL methods use language-oriented adaptation modules to adapt the transformer model and lack the image-related (multi-scale) local priors for visual tasks (especially for VPR). This raises the need to develop a new adaptation method to introduce multi-scale local priors to the foundation model for VPR.

In this paper, we propose a novel method to learn \textbf{Cr}oss-\textbf{i}mage \textbf{c}orrelation-\textbf{a}ware representation for VPR, abbreviated as CricaVPR. Our method first uses a backbone with a pooling module to yield initial feature representations. Then we use a cross-image encoder equipped with the attention to calculate the correlation between multiple image representations within a batch to get final features. The images in a batch can be taken in the same place under different conditions (e.g. lighting) or from different viewpoints, or even captured from different places. This process allows each feature to enhance itself with useful information from others, thus producing condition-robust, viewpoint-robust, and discriminative representations. Meanwhile, we use the visual foundation model as the backbone in our architecture, and design a multi-scale convolution-enhanced adaptation method, in which we freeze the pre-trained foundation models and insert a few trainable lightweight adapters, to introduce the multi-scale local prior knowledge and adapt the foundation model for the VPR task. 

Our work brings the following \textbf{contributions}: \textbf{1)} We propose a cross-image correlation-aware representation method, which uses the attention mechanism to model the correlation between multiple image representations within a batch and make each feature more robust. \textbf{2)} We design a parameter-efficient adaptation method to adapt pre-trained models for VPR, in which the proposed multi-scale convolution adapter is used to introduce multi-scale local information to boost performance. \textbf{3)} Extensive experiments on the benchmark datasets show that our method can outperform the state-of-the-art (SOTA) methods by a large margin with less training time. The results on Pitts30k that best reflect the advantages of our method are shown in Fig. \ref{fig:intro}.

\section{Related Work}
\textbf{Visual Place Recognition:}
The early VPR approaches typically represent place images by global features that are computed using aggregation algorithms, such as Bag of Words \cite{BoW} and Vector of Locally Aggregated Descriptors (VLAD) \cite{vlad2010,VLAD1,allVLAD,densevlad,pbVLAD}, to aggregate the hand-crafted descriptors like SURF \cite{SURF,fab08}. Then these methods perform a nearest neighbor search in the global feature space over the database to get the most similar images. With the significant success of deep learning on various computer vision tasks, most recently VPR methods \cite{sunderhaufIROS2015, netvlad, crn, SPED, landmarks2, semantic,categorization1, categorization2, landmarks3, yin2019, gcl, gsv, mixvpr,sfrs, cosplace, eigenplaces} 
have employed a variety of deep features to represent place images for boosting performance. Likewise, the aggregation algorithm has also been changed into a differentiable module to embed neural networks for end-to-end training \cite{netvlad,DBOW,attentionVLAD}.  
However, most of the global-retrieval-based methods lack robustness in challenging environments and are prone to perceptual aliasing.

Two typical ways to alleviate this issue are to impose temporal consistency constraints and spatial consistency constraints. The former performs image sequence matching (i.e. utilize temporal continuity) \cite{seqslam,hmmpr,hmmpr2,sta-vpr,seqnet} to realize robust VPR in challenging environments. The latter is often developed as a two-stage VPR system \cite{hvpr,patchvlad,geowarp,aanet,transvpr,r2former,dhevpr,selavpr}, which searches for top-k candidate images over the database using global features, then performs spatial consistency matching using local features to re-rank candidates. Different from these methods bringing additional constraints, runtime, and memory overhead, our model learns highly robust global representation via cross-image correlation awareness for global-retrieval-based VPR.

\textbf{Parameter-efficient Transfer Learning:} Some recent studies \cite{clip,florence,wang2022image,dinov2} trained the large transformer-based foundation models on huge quantities of data. These models are capable of producing well-generalized feature representation and performing admirably on some common visual tasks. A promising technique for adapting these foundation models to more diverse downstream tasks with only fine-tuning a few (extra) parameters is PETL \cite{adapter,prompttuning,lora}, which is initially proposed in natural language processing to address the catastrophic forgetting issue \cite{catastrophic} and reduce training costs. Training the inserted task-specific adapters \cite{adapter} while keeping the pre-trained foundation models frozen is one of the commonly used PETL methods, and we follow it in our work. There are multiple adapter-based methods \cite{sideadapter,convpass,adaptformer,Stadapter,contrastAlign,dualadapt,aim} have been proposed to address a wide range of visual tasks. A closely related work to ours is Convpass \cite{convpass}, which used convolutional bypasses in ViT as adaptation modules to introduce image-related local inductive biases and avoid performance degradation in downstream fine-tuning. However, our work designs a multi-scale convolution adapter to learn more proper local information to improve the performance on the VPR task.

\section{Methodology}
Our method involves the Vision Transformer (ViT) and the attention mechanism used in it. So we first briefly review them in this section. Then we propose the cross-image correlation-aware representation method to describe place images. Finally, we present the multi-scale convolution-enhanced adaptation method to adapt the foundation model for VPR and the training strategy for fine-tuning.

\subsection{Preliminary}
The ViT model \cite{vit} and its variants have been applied for many computer vision tasks including VPR mainly due to its superior performance in modeling long-range dependencies. To process an input image with ViT, the image is initially divided into $N$ non-overlapping patches, which are then linearly projected into $D$-dim patch embeddings $x_p\in \mathcal{R}^{N \times D}$. Meanwhile, a learnable \verb|[class]| token is prepended to $x_p$ to form $x_0=[x_{class};x_p]\in \mathcal{R}^{(N+1) \times D}$. To preserve the original positional information of each patch token, the corresponding positional embeddings are added to $x_0$ to get $z_0$, which is fed into a series of transformer encoder layers to yield the feature representation. 
A transformer encoder layer consists of three main components: the multi-head attention (MHA) layer, the MLP layer, and the LayerNormalization (LN) layer. The forward process of input $z_{l-1}$ passing through a transformer encoder layer to yield the output $z_l$ can be formulated as
\begin{equation}
	\begin{aligned}
		z^{\prime}_l&=\text{MHA}\left(\text{LN}\left(z_{l-1}\right)\right)+z_{l-1},\\
		z_l&=\text{MLP}\left(\text{LN}\left(z^{\prime}_l\right)\right)+z^{\prime}_l.
	\end{aligned}
	\label{eq:encoder}
\end{equation}

The MLP layer is made up of two fully connected layers, which are mainly used for feature nonlinearization and dimension conversion. Here we briefly overview the process of calculating the correlation and attention in the MHA layer. The input sequence is first linearly transformed to produce the queries $Q$, keys $K$, and values $V$. Then the attention among $Q, K$ and $V$ is computed using the Scaled Dot-Product Attention \cite{transformer}, denoted as
\begin{equation}
	\mathit { Attn }(Q, K, V)=\mathit{Softmax}\left(Q K^{\top} / \sqrt{d}\right) V.
\end{equation}
The MHA utilizes different learnable linear projections to generate the queries, keys, and values $h$ times and performs attention for each set of projections in parallel. Specifically, we first compute the attention scores between each query and all keys, establishing the correlations between them. These scores are then multiplied with the corresponding values to model dependencies among these tokens. Finally, the outputs of $h$ attention heads are concatenated (and once again projected). All tokens/elements in the input sequence are correlated in this process. In the next section, we will also use this attention mechanism to compute the across-image correlation.

There are two ways to yield global representations of places using the output of ViT. The first is to directly use the output class token as a global feature. The second is to reshape the output patch tokens as a feature map (similar to the output of CNN) to restore the spatial position, and use the aggregation/pooling method (e.g. GeM \cite{gem}) to process it as a global feature. Both the class token and GeM pooling are used to produce the place representation in our work.
 
\subsection{Cross-image Correlation-aware Place Representation}
 \begin{figure}[!t]
	\centering
	\includegraphics[width=0.85\linewidth]{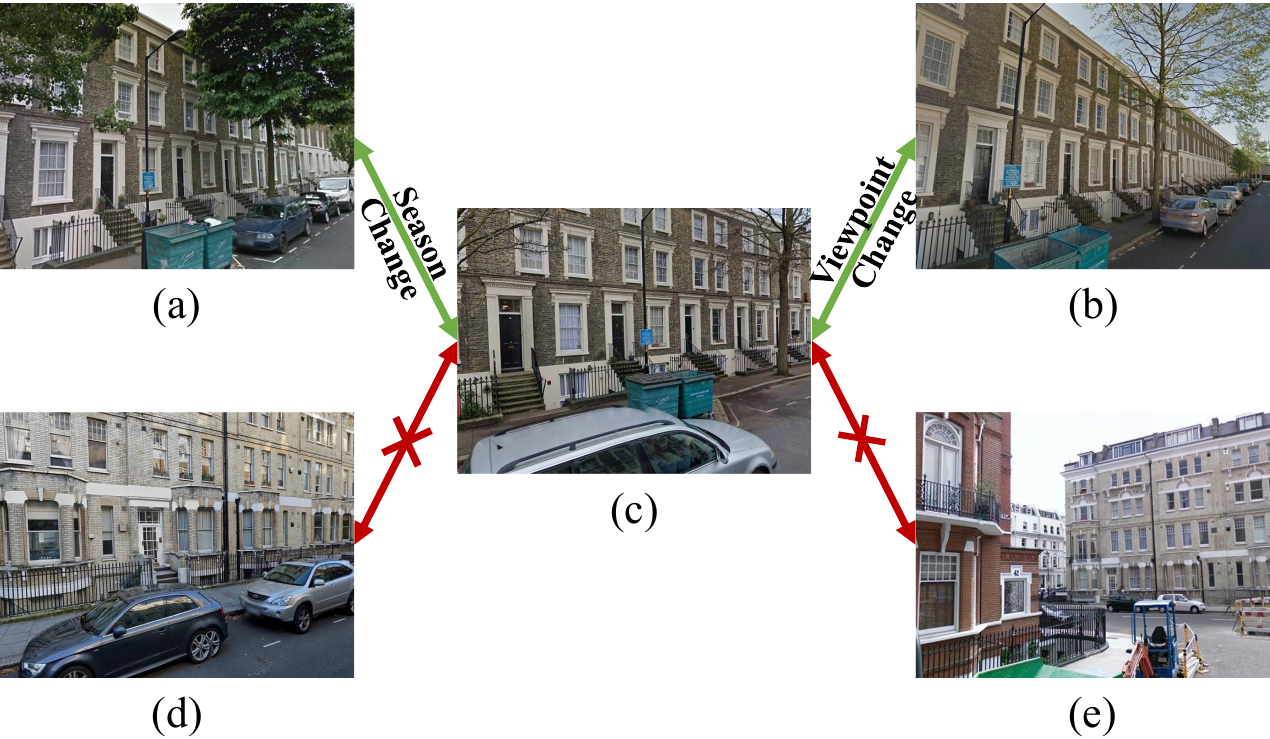}
    \vspace{-0.3cm}
	\caption{
		\textbf{The example of partial images in a batch.} (a), (b), and (c) are taken from the same place with different conditions (seasons) and viewpoints. (d), (e), and (c) are captured from different places, but (d) is similar to (c). When the model produces the features of (c), it can harvest relevant information from other images to yield a better representation.
	}
    \vspace{-0.5cm}
	\label{fig:motivation}
\end{figure}

\begin{figure*}[!t]
	\centering	\includegraphics[width=0.92\linewidth]{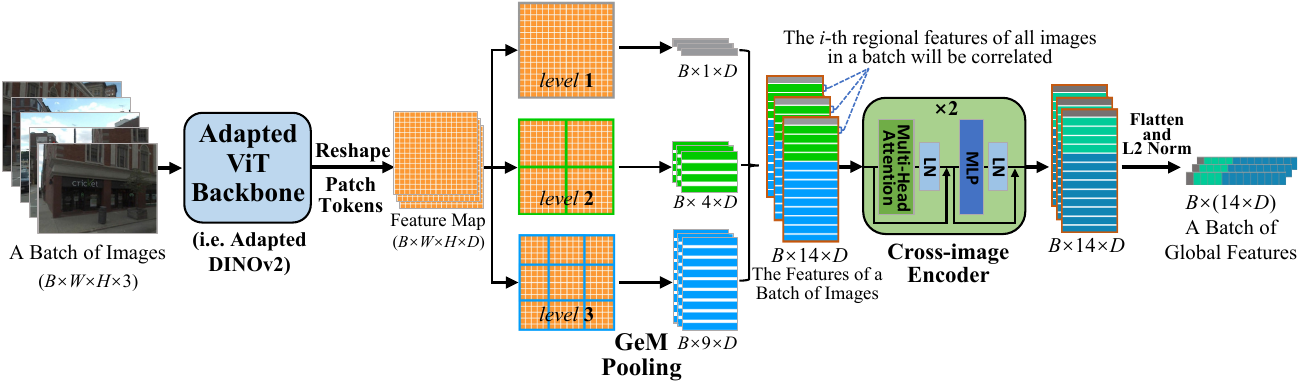}
    \vspace{-0.3cm}
	\caption{
		\textbf{The pipeline to produce the proposed cross-image correlation-aware representation.} The cross-image encoder is the core component for modeling correlations between different image features in a batch. Note that we are correlating the $i$-th regional features of all images in a batch, not all regional features of an image. Besides, the cross-image encoder consists of 2 stacked vanilla transformer encoder layers \cite{transformer} with the LN layer behind the MHA/MLP layer, which is different from that in ViT \cite{vit} (LN is before MHA/MLP).
	}
    \vspace{-0.4cm}
	\label{fig:archit}
\end{figure*}
The methods based on neural networks have dominated the VPR area over the past decade. These methods commonly produce the deep feature representation of an image with only this image itself. Such features often lack robustness in challenging environments and are incapable of addressing the perceptual aliasing issue. In this work, we present a simple and effective solution to this problem. We attempt to correlate the features of place images in a batch, so that each image representation can harvest useful information from the other image representations to enhance its own robustness. More specifically, there may be images from the same place but taken from different viewpoints or under different conditions, or images from different places that look similar (or not) in a batch, as shown in Fig. \ref{fig:motivation}. On the one hand, image representations from the same place with different perspectives and conditions can improve the viewpoint invariance and condition invariance of each other after the correlated encoding. On the other hand, image representations from different places also promote each other to produce discriminative features. As a result, our method can provide highly robust global representations to deal with viewpoint changes, condition changes, and perceptual aliasing.

We design the pipeline to produce desired global features as shown in Fig. \ref{fig:archit}. The output patch tokens of a batch of images from the ViT backbone are reshaped as the $B\times W\times H\times D$-dim (i.e., $batch$ $size\times weight\times height\times token$ $dimension$) feature maps. We first use the spatial pyramid \cite{spm} to produce initial feature representations. The feature maps are split at three levels (1$\times$1, 2$\times$2, and 3$\times$3). Then we use GeM pooling to process local (patch) features within the divided regions and get a total of 14 regional features of each image. Since the first level is a global aggregation, we directly use the class token to replace the GeM feature in this level for better performance. Next is the most critical step. We treat the $i$-th regional features of all images in a batch as a sequence of embedding vectors $f_i$, that is
\begin{equation}
	f_i=\{f_i^1,f_i^2,...,f_i^B\}  \quad i\in\{1,2,...,14\}, 
\end{equation}
and feed the 14 sequences of embedding vectors into a cross-image encoder to model the correlation between the $i$-th regional features of all images in a batch. That is, we apply the cross-image encoder to process each $f_i$ to correlate images in a batch. Instead of directly using the attention (MHA) layer, the cross-image encoder is structured using two (vanilla) transformer encoder layers \cite{transformer} that also include the MLP layer, LN layer, and skip connection for stable training and better performance. The 14 output regional features of each image are sequentially concatenated (i.e., flattened) and L2-normalized as the final global representation of the image.

It should be noted that the retrieval process of our method is the same as the common global-retrieval-based method. That is, it uses the global feature of a single image for retrieval. Besides, we choose the spatial pyramid to produce the initial feature in order to subsequently correlate images at different scales, and the final sequential concatenation of regional features also introduces spatial position information to the global representations. In fact, our method can also use other aggregation methods to yield initial features, and boost the performance of these methods.

\subsection{Multi-scale Convolution-enhanced Adaptation}
Our work adapts the distilled DINOv2 \cite{dinov2} as the backbone (i.e. the adapted DINOv2/ViT backbone in Fig. \ref{fig:archit}), which is based on ViT-B/14. The DINOv2 work trains the ViT model on the large-scale curated LVD-142M dataset with the self-supervised strategy, and can provide powerful visual features to achieve promising performance on some common tasks without any fine-tuning. AnyLoc \cite{anyloc} is a VPR work that uses pre-trained DINOv2 without fine-tuning. However, there exists a gap between the tasks of model pre-training and VPR due to the inherent difference in training objectives and data. Directly using such a pre-trained model in VPR cannot fully unleash its powerful capability.

\begin{figure}[!t]
	\centering
	\includegraphics[width=0.95\linewidth]{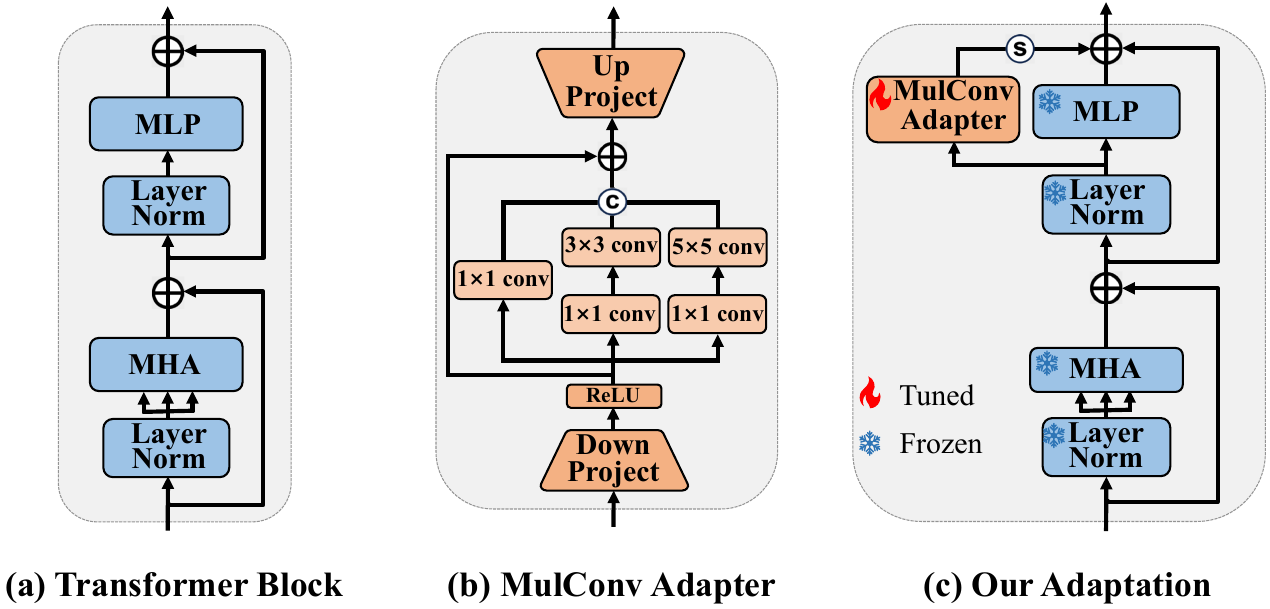}
	\vspace{-0.3cm}
	\caption{
		\textbf{Illustration of our multi-scale convolution-enhanced adaptation.} (a) is a transformer block in ViT. (b) is the MulConv adapter. We add the MulConv adapter in parallel to the MLP layer in each transformer block to achieve our adaptation as in (c).
	}
	\vspace{-0.4cm}
	\label{fig:adapter}
\end{figure}

The adapter-based parameter-efficient transfer learning \cite{adapter} provides an effective way to adapt foundation models for downstream tasks, which freezes the pre-trained model and only fine-tunes the added lightweight adapter. The vanilla adapter is a bottleneck module consisting of a down-projection (fully connection) layer, an up-projection layer, and a non-linearity (activation) layer in the middle. The Convpass work \cite{convpass} applies convolution layers to introduce image-related local inductive biases into models. However, we found that improper local priors provided by Convpass risk reducing performance in VPR. Inspired by the inception module in GoogLeNet \cite{inception}, we design our multi-scale convolution (MulConv) adapter as shown in Fig. \ref{fig:adapter} (b).

Different from the vanilla adapter, our MulConv adapter adds a MulConv module between the (ReLU) activation layer and the up-projection layer. This module consists of three parallel convolutional paths of different scales (1$\times$1, 3$\times$3, 5$\times$5). The 1$\times$1 convolution is also used before the 3$\times$3 and 5$\times$5 convolutions to reduce channel dimension. This design and the bottleneck structure of the adapter make our MulConv adapter still lightweight. The outputs of the three convolutional paths are concatenated to form the output of the MulConv module. Besides, there is a skip connection in parallel to the MulConv module. Finally, the MulConv adapter is added in parallel to the MLP layer (multiplied by a scaling factor $s$) in each transformer block (i.e. transformer encoder layer) of the ViT backbone to achieve multi-scale convolution-enhanced adaptation, which can introduce proper (multi-scale) local priors to the model and improve performance for VPR. So the computation of each adapted transformer block can be denoted as
\begin{equation}
	\begin{aligned}		z^{\prime}_l&=\text{MHA}\left(\text{LN}\left(z_{l-1}\right)\right)+z_{l-1},\\
z_l&=\text{MLP}\left(\text{LN}\left(z^{\prime}_l\right)\right)+s\cdot\text{Adapter}\left(\text{LN}\left(z^{\prime}_l\right)\right)+z^{\prime}_l.
	\end{aligned}
	\label{eq:adapter}
\end{equation}

\subsection{Training Strategy} 
We train our model on the GSV-Cities \cite{gsv} dataset with full supervision. This dataset contains 560k images captured at 67k places with highly accurate labels. We follow the standard framework of this dataset and use the multi-similarity (MS) loss \cite{multiloss} with online hard mining strategy for training. The MS loss is computed as
\begin{equation}
\begin{split}
	\label{eq:MS}
	\mathcal{L}_{MS} = \frac{1}{B}\sum_{q=1}^B  \bigg\{\frac{1}{\alpha}  { \log \big[1 + \sum_{p  \in \mathcal{P}_q } e^{-\alpha (S_{qp} - \lambda)}}\big]  \\
	+ \frac{1}{\beta }  { \log \big[1+ \sum_{n \in \mathcal{N}_q}
		 e^{\beta (S_{qn} - \lambda)} \big]} \bigg\},
\end{split}   
\end{equation}
where for each query (anchor) image $I_q$ in a batch, $\mathcal{P}_q$ is the set of indices $\{p\}$ that correspond to the positive samples for $I_q$, and $\mathcal{N}_q$ is the set of indices $\{n\}$ that correspond to the negative samples for $I_q$. $S_{qp}$ and $S_{qn}$ are the cosine similarities of a positive pair $\{I_q,I_p\}$ and a negative pair $\{I_q,I_n\}$.
$\alpha$, $\beta$ and $\lambda$ are three set constants (hyperparameters).

\section{Experiments}
\subsection{Datasets and Performance Evaluation}
\begin{table}
  \begin{center}
  \scalebox{0.8}{ 
  \renewcommand{\arraystretch}{1}
  \begin{tabular}{cccc}
    \toprule
    \multirow{2}{*}{Dataset} & \multicolumn{1}{c}{\multirow{2}{*}{Description}} & \multicolumn{2}{c}{Number}  \\ \cline{3-4} 
     \multicolumn{1}{c}{}& & Database & Queries \\ 
     \midrule
    Pitts30k & urban, panorama & 10,000 & 6,816  \\ \hline
    MSLS-val &  urban, suburban & 18,871 & 740  \\
    MSLS-challenge & long-term & 38,770 & 27,092  \\ \hline
    Tokyo24/7 & urban, day/night & 75,984 &  315 \\ \hline
    Nordland & natural, seasonal & 27,592 &  27,592 \\\hline
    SVOX & cross-domain & 17,166 &  4,356 \\ \hline
    AmsterTime & very long-term & 1,231 &  1,231 \\
    \bottomrule
  \end{tabular}}
  \vspace{-0.6cm}
\end{center}
\caption{Summary of the test datasets in experiments.}
\vspace{-0.6cm}
\label{tab:datasets}
\end{table}

The experiments are conducted on several VPR benchmark datasets. These datasets exhibit viewpoint changes, condition changes, and the perceptual aliasing issue. Table \ref{tab:datasets} summarizes the key information of them. \textbf{Pitts30k} \cite{pitts} mainly shows large viewpoint changes. \textbf{MSLS} \cite{msls} consists of images captured in urban, suburban, and natural scenes over 7 years, and covers various visual changes. \textbf{Tokyo24/7} \cite{densevlad} exhibits severe illumination (day/night) changes. We also use three challenging datasets: \textbf{Nordland} (with seasonal changes) \cite{eigenplaces}, \textbf{SVOX} (cross-domain dataset) \cite{SVOX}, and \textbf{AmsterTime} (with very long-term changes) \cite{amstertime}. More details are in Supplementary (Suppl.) Material.

The Recall@N (R@N) metric is used in our experiments to evaluate recognition performance. It is the percentage of queries for which at least one of the N retrieved database images is taken within a threshold of ground truth. We set the threshold to 25 meters and 40$^{\circ}$ for MSLS, 25 meters for Pitts30k, Tokyo24/7, and SVOX, $\pm 10$ frames for Nordland, unique counterpart for AmsterTime, following common evaluation procedures \cite{msls,pitts,densevlad}.

\begin{table*}
  \centering
  \setlength{\tabcolsep}{1.2mm}{
  \begin{tabular}{@{}l|c||ccc||ccc||ccc||ccc}
  \toprule
\multirow{2}{*}{Method} & \multirow{2}{*}{Dim} & \multicolumn{3}{c||}{Pitts30k} & \multicolumn{3}{c||}{Tokyo24/7} & \multicolumn{3}{c||}{MSLS-val} & \multicolumn{3}{c}{MSLS-challenge} \\
\cline{3-14}
& & R@1 & R@5 & R@10 & R@1 & R@5 & R@10 & R@1 & R@5 & R@10  & R@1 & R@5 & R@10 \\
\hline
NetVLAD \cite{netvlad} &32768 & 81.9 &91.2 &93.7 & 60.6 & 68.9 & 74.6 & 53.1 &66.5 &71.1 &35.1 & 47.4 & 51.7  \\
SFRS \cite{sfrs} &4096 & 89.4 & 94.7 & 95.9 & 81.0 & 88.3 & 92.4 & 69.2 & 80.3 & 83.1 & 41.6 & 52.0 &56.3  \\
Patch-NetVLAD \cite{patchvlad} &/ & 88.7 & 94.5 & 95.9 & \underline{86.0} & 88.6 & 90.5 & 79.5 & 86.2 & 87.7 & 48.1 &57.6 & 60.5  \\
TransVPR \cite{transvpr} &/ &  89.0 &  94.9 &  96.2 & 79.0 &  82.2 & 85.1 & 86.8 & 91.2 & 92.4 & 63.9 & 74.0  & 77.5   \\
CosPlace \cite{cosplace} &512 & 88.4 & 94.5 & 95.7 & 81.9 & 90.2 & 92.7 & 82.8 & 89.7 & 92.0  & 61.4 & 72.0 & 76.6 \\
GCL \cite{gcl} &2048  &80.7 &91.5 &93.9 &69.5 &81.0 &85.1  &79.5 &88.1 &90.1 &57.9 &70.7 &75.7 \\
MixVPR \cite{mixvpr} &4096 & 91.5 & 95.5 & 96.3 & 85.1 & 91.7 & 94.3 & 88.0 & 92.7 & 94.6 & 64.0 & 75.9 & 80.6 \\
EigenPlaces \cite{eigenplaces} &2048 &\underline{92.5} &\underline{96.8} &\underline{97.6} &\textbf{93.0} &\underline{96.2} &\underline{97.5} &\underline{89.1} &\underline{93.8} &\underline{95.0}  &\underline{67.4} &\underline{77.1} &\underline{81.7}  \\
\hline
CricaVPR (ours) &4096 & \textbf{94.9} & \textbf{97.3} & \textbf{98.2} & \textbf{93.0} & \textbf{97.5} & \textbf{98.1} &  \textbf{90.0} &  \textbf{95.4} & \textbf{96.4} &  \textbf{69.0} &  \textbf{82.1} &  \textbf{85.7}  \\
\bottomrule
\end{tabular}}
\vspace{-0.2cm}
\caption{\textbf{Comparison to state-of-the-art methods on benchmark datasets.} The best is highlighted in \textbf{bold} and the second is \underline{underlined}.}
\vspace{-0.25cm}
\label{tab:compare_SOTA}
\end{table*}

\begin{figure*}[!t]
	\centering
	\includegraphics[width=0.86\linewidth]{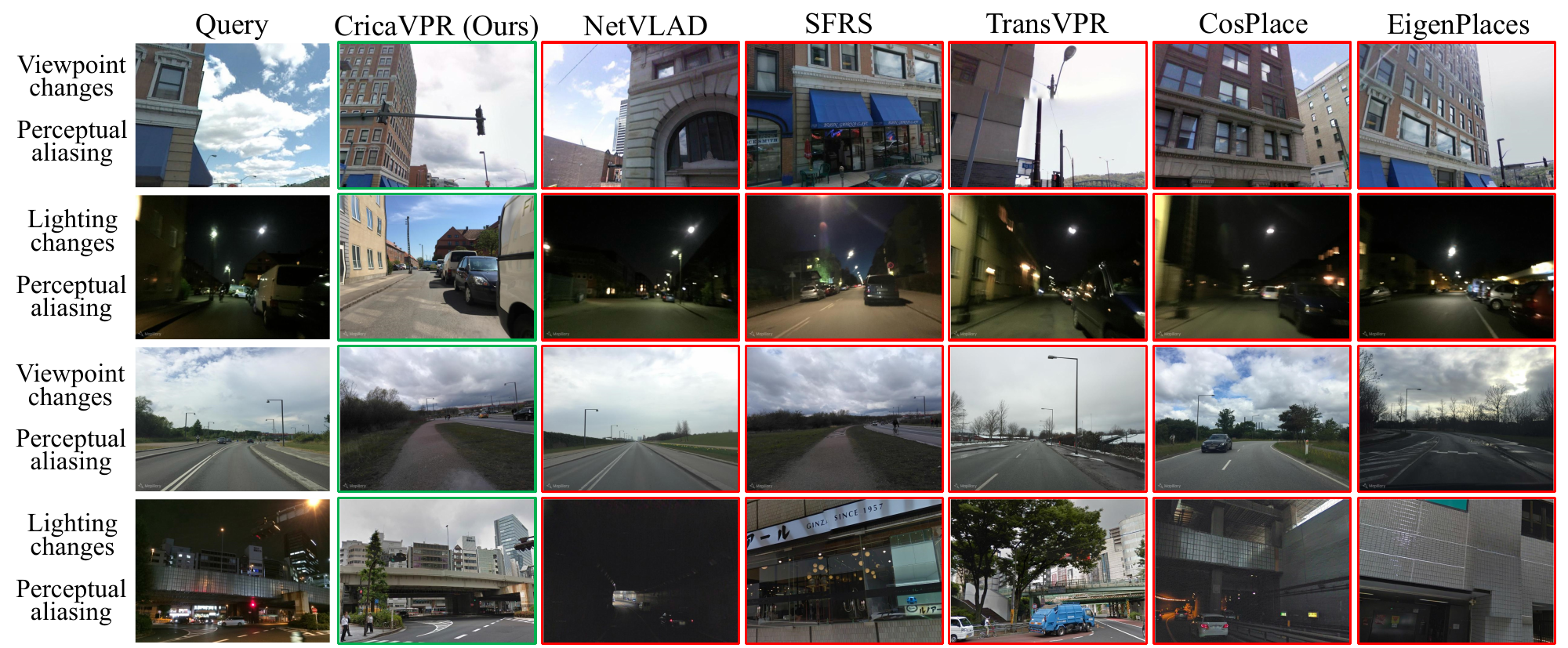}
	\vspace{-0.23cm}
	\caption{
		\textbf{Qualitative results.} These four challenging examples show severe viewpoint changes and condition changes. The proposed CricaVPR successfully yields the right results, while other methods return incorrect images. In each example, there are methods to return similar images from different places (i.e., incorrect) due to perceptual aliasing. In the second example, the query image is taken at night, causing all the other methods to return night images but from different places (i.e. wrong). However, our method returns an image taken during the day at the same place (i.e. correct).
	}
	\vspace{-0.43cm}
	\label{fig:result}
\end{figure*}

\begin{table}
  \centering
  \setlength{\tabcolsep}{1.5mm}{
  \renewcommand{\arraystretch}{0.98}
  \begin{tabular}{@{}c|cccc}
  \toprule
\multirow{2}{*}{Method}   & \multirow{2}{*}{Nordland} & Amster & SVOX & SVOX 
\vspace{-0.05cm}
\\
& & Time & -Night & -Rain \\
\hline
SFRS \cite{sfrs} & 16.0 & 29.7  & 28.6 & 69.7 \\
CosPlace \cite{cosplace} & 58.5 & 38.7  & 44.8 & 85.2 \\
MixVPR \cite{mixvpr} & \underline{76.2} & 40.2  & \underline{64.4} & \underline{91.5} \\
EigenPlaces \cite{eigenplaces} & 71.2 & \underline{48.9} & 58.9 & 90.0 \\
\hline
CricaVPR (ours)  & \textbf{90.7} & \textbf{64.7} & \textbf{85.1} & \textbf{95.0} \\
\bottomrule
\end{tabular}}
\vspace{-0.2cm}
\caption{\textbf{Comparison (R@1) to SOTA methods on more challenging datasets.} More results are in Suppl. Material.}
\label{tab:compare_SOTA2}
\vspace{-0.45cm}
\end{table}

\subsection{Implementation Details}
We fine-tune our model on two NVIDIA GeForce RTX 3090 GPUs using PyTorch. The resolution of the input image is 224×224 and the token dimension of the backbone (ViT-B/14) is 768. Our model outputs the 14×768-dim original global features, and we use PCA for dimensionality reduction. The bottleneck ratio of our adapters is set to 0.5, so the input dimension of the three convolutional paths is 384. The 1×1 convolution before the 3×3 and 5×5 convolution reduces the channels to 24. The output dimensions of the three convolutional paths are 192, 96, and 96. The scaling factor $s$ in Eq. \ref{eq:adapter} is set to 0.2. We set the hyperparameters $\alpha=1, \beta=50, \lambda=0$ in Eq. \ref{eq:MS} and margin = 0.1 in online mining, as in GSV-Cities \cite{gsv}. We fine-tune the model using the Adam optimizer with the initial learning rate set as 0.0001 and multiplied by 0.5 after every 3 epochs. A training batch contains 72 places with 4 images each (i.e. 288 images). Training is implemented until the R@5 on Pitts30k does not improve for 3 epochs. An inference batch contains 8 images for Pitts30k and 16 images for others.

\subsection{Comparison with State-of-the-Art Methods}
In this section, we compare our CricaVPR with several SOTA VPR methods, mainly including six global-retrieval-based methods: NetVLAD \citep{netvlad}, SFRS \citep{sfrs}, CosPlace \citep{cosplace}, GCL \cite{gcl}, MixVPR \citep{mixvpr} and EigenPlaces \cite{eigenplaces}. Note that our work uses the same training dataset as MixVPR, i.e., GSV-Cities. Meanwhile, CosPlace and EigenPlaces are trained on individually constructed extra large-scale datasets, i.e., SF-XL. Both MixVPR and EigenPlaces are the latest works and represent the SOTA performance of the VPR methods based on global feature retrieval. Additionally, we also compare our approach with two excellent two-stage VPR methods (Patch-NetVLAD \cite{patchvlad} and TransVPR \cite{transvpr}), which require time-consuming re-ranking using local features. The details of these methods are in Suppl. Material. Table \ref{tab:compare_SOTA} shows the quantitative results on Pitts30k, Tokyo24/7, and MSLS. Our CricaVPR uses PCA to reduce the feature dimensionality to 4096-dim (in this subsection), and achieves the best R@1/R@5/R@10 on all datasets.

MixVPR, EigenPlaces, and our CricaVPR all achieve excellent performance on these datasets. Especially on Pitts30k, which shows significant viewpoint changes but no drastic condition changes, EigenPlaces achieves 92.5\% R@1. This indicates that the challenge posed by viewpoint changes has been effectively addressed by existing methods (i.e., EigenPlaces and MixVPR). However, our method continues to improve performance on Pitts30k, achieving an impressive 94.9\% R@1. This improvement primarily stems from the powerful ability of our method to produce more discriminative global representations to differentiate similar images from different places, i.e., address perceptual aliasing. The MSLS dataset is more challenging as it shows severe condition variations and includes some suburban or natural scene images lacking landmarks and prone to perceptual aliasing. Nevertheless, our method achieves 95.4\% R@5 on MSLS-val and 82.1\% R@5 on MSLS-challenge, showing significant advantages over other global-retrieval-based methods and two-stage methods.

Fig. \ref{fig:result} qualitatively demonstrates the superior performance of our method in some extreme environments. These challenging examples include drastic condition changes, viewpoint changes, or only small regions in the images showing discriminative objects. In these examples, other methods either get similar images but from different places (i.e. suffer from perceptual aliasing), or retrieve places that are close in geographical distance but still out of the set threshold, that is, they fail to retrieve the correct results. Our approach shows high robustness against these challenges.

To further evaluate the performance of our method in extreme scenarios, we conduct experiments on three challenging datasets: Nordland, which exhibits seasonal changes; AmsterTime, which spans a very long time period; and SVOX, which shows extreme illumination and weather variations. The results, as shown in Table \ref{tab:compare_SOTA2}, demonstrate the significant superiority of our method compared to other SOTA methods. Our CricaVPR outperforms all other SOTA methods with 14.5\%, 15.8\%, and 20.7\% absolute R@1 improvements on Nordland, AmsterTime, and SVOX-Night, respectively. This further highlights that the global image representation of our method is highly robust.

\subsection{Ablation Study}
We perform a series of ablation experiments to validate the effectiveness of the proposed components in our method. All ablated methods no longer use PCA for dimensionality reduction by default. We will conduct separate experiments to show the impact of feature dimensions on the results.

\begin{table}
  \centering
  \small
  \setlength{\tabcolsep}{1mm}{
  \renewcommand{\arraystretch}{0.92}
  \begin{tabular}{@{}c||cc||cc||cc}
  \toprule
\multirow{2}{*}{Ablated versions}   & \multicolumn{2}{c||}{Pitts30k} & \multicolumn{2}{c||}{Tokyo24/7} & \multicolumn{2}{c}{MSLS-val} \\
\cline{2-7}
& R@1 & R@5 & R@1 & R@5 & R@1 & R@5 \\
\hline
FrozenDINOv2-SPM & 74.8 & 90.1  & 49.8 & 67.0 & 45.4 & 60.7  \\
\hline
AdaptGeM & 87.1 & 94.0 & 70.2 & 85.4 & 78.4 & 87.8 \\
AdaptSPMG & 87.8 & 94.1 & 72.1 & 85.1 & 78.0 & 88.4  \\
AdaptSPM & 90.6 & 95.9 & 85.1  & 93.3 & 85.5 & 93.2 \\
\hline
AdaptGeM\textbf{+Crica}  & 93.9 & 97.2 & 87.6 & 93.3 & 86.1 & 93.4 \\
AdaptSPMG\textbf{+Crica}  & 94.3 & 97.3 & \textbf{93.7} & 96.5 & 89.7 & 95.3  \\
AdaptSPM\textbf{+Crica}  & \textbf{94.8} & \textbf{97.4} & 93.0 & \textbf{97.1} & \textbf{89.9} & \textbf{95.4} \\
\bottomrule
\end{tabular}}
\vspace{-0.15cm}
\caption{\textbf{Ablation on cross-image awareness.} The ``\textbf{+Crica}" represents the addition of our cross-image correlation awareness to get the final global feature. The ``SPM" represents our spatial pyramid model representation, while ``SPMG" is the spatial pyramid model solely based on GeM. Except for the FrozenDINOv2-SPM that directly uses an untuned backbone (as baseline), all other versions use our adaptation method for fine-tuning.}
\vspace{-0.5cm}
\label{tab:ablation_Crica}
\end{table}

\textbf{Ablation on cross-image correlation awareness.} 
The cross-image correlation awareness achieved by the cross-image encoder after the backbone is the most important module in our method. We compare the performance of the three kinds of global features with or without across-image awareness. These features are GeM, the spatial pyramid model representation solely based on GeM (SPMG), and our spatial pyramid model representation (SPM) which uses both class token and GeM. The results are shown in Table \ref{tab:ablation_Crica}. After incorporating the proposed cross-image correlation awareness (Crica), all three features achieve significant performance improvements. Due to the already impressive performance of the SPM feature after our model adaptation (AdaptSPM), the improvement provided by Crica on this feature is not as pronounced as on the GeM and SPMG features. Nevertheless, AdaptSPM+Crica still resulted in 4.2\%, 7.9\%, and 4.4\% absolute R@1 improvements over the AdaptSPM feature on Pitts30k, Tokyo24/7, and MSLS-val, respectively. Moreover, AdaptGeM+Crica achieves an impressive 17.4\% absolute R@1 improvement over AdaptGeM on Tokyo24/7. With the combined effect of our Crica and model adaptation, our method achieves nearly 2× higher R@1 on Tokyo24/7 and MSLS-val compared to the direct use of frozen DINOv2 with the SPM representation (FrozenDINOv2-SPM).

\begin{figure}[!t]
	\centering	\includegraphics[width=0.98\linewidth]{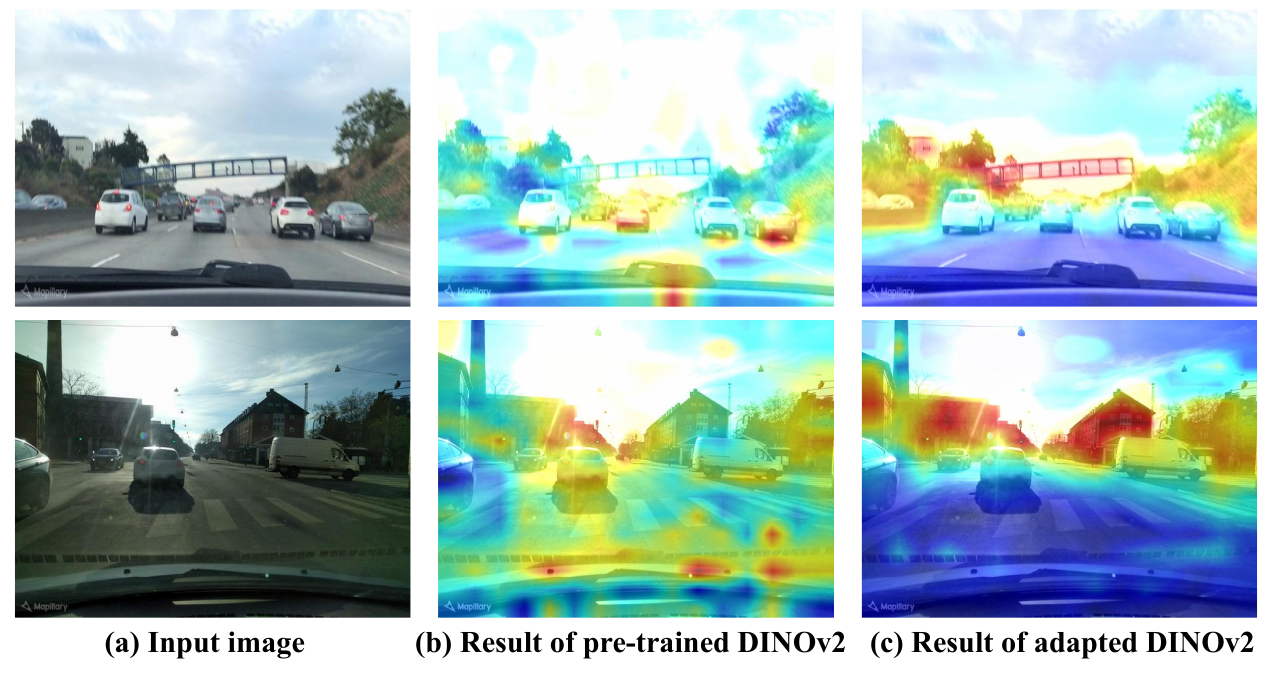}
    \vspace{-0.3cm}
    \caption{
		\textbf{The output feature map (attention) visualizations} of pre-trained DINOv2 and adapted DINOv2. The regions attended to by pre-trained DINOv2 have no relevance to place recognition. However, adapted DINOv2 focuses on discriminative areas for VPR. Buildings that remain relatively unchanged over time receive the highest attention. Vegetation that is not expected to change in the short term receives moderate attention. Non-discriminative elements such as the sky, ground, and dynamic vehicles, are ignored.
	}
    \vspace{-0.5cm}
	\label{fig:attention}
\end{figure}
\textbf{Ablation on adaptation.}
We first use only the GeM features alone (without cross-image awareness) to demonstrate the performance improvement achieved by our adaptation method. As shown in Table \ref{tab:ablation_adapt}, MulConvAdapter-GeM using our adaptation achieves a significant improvement over FrozenDINOv2-GeM. Especially on MSLS-val, which has more dynamic interference, our adaptation achieves nearly 2× higher R@1. Fig. \ref{fig:attention} vividly illustrates the underlying reasons. The adapted DINOv2, in contrast to the pre-trained DINOv2, exhibits a stronger ability to focus on objects related to place recognition, with more attention given to more important objects. Table \ref{tab:ablation_adapt} also shows the performance of different fine-tuning methods when using the proposed global features (i.e. the SPM feature with cross-image correlation awareness). The FullTunedDINOv2 achieves a notable improvement over FrozenDINOv2 on Pitts30k and MSLS-val. However, because our training data has no night images like those in Tokyo24/7, FullTunedDINOv2 performs worse than FrozenDINOv2 on Tokyo24/7, i.e., it suffers from catastrophic forgetting. This indicates the necessity of parameter-efficient fine-tuning (using an adapter). Besides, ConvAdapter (as in Convpass \cite{convpass}) uses 3×3 convolution to introduce local inductive biases into the model. However, it brings inappropriate local priors for VPR and results in performance degradation compared to VanillaAdapter. Our method (MulConvAdapter) uses multi-scale convolution to introduce more proper local information and thus achieves the best performance.

\begin{table}
  \centering
  \small
  \setlength{\tabcolsep}{0.8mm}{
  \renewcommand{\arraystretch}{0.92}
  \begin{tabular}{@{}c||cc||cc||cc}
  \toprule
\multirow{2}{*}{Ablated versions}   & \multicolumn{2}{c||}{Pitts30k} & \multicolumn{2}{c||}{Tokyo24/7} & \multicolumn{2}{c}{MSLS-val} \\
\cline{2-7}
& R@1 & R@5 & R@1 & R@5 & R@1 & R@5 \\
\hline
FrozenDINOv2-GeM & 79.2 & 90.1  & 65.4 & 83.8 & 40.8 & 51.5  \\
MulConvAdapter-GeM & 87.1 & 94.0 & 70.2 & 85.4 & 78.4 & 87.8 \\
\hline
FrozenDINOv2 & 79.2 & 90.1  & 80.0 & 89.8 & 58.8 & 71.2  \\
FullTunedDINOv2 & 94.1 & 96.6  & 76.8 & 88.3 & 86.2 & 93.2  \\
VanillaAdapter & 94.6 & \textbf{97.4} & 92.7 & 96.5 & 89.2 & \textbf{95.5} \\
ConvAdapter  & 93.8 & 96.9 & 92.7 & 95.9 & 88.0 & 94.2 \\
MulConvAdapter  & \textbf{94.8} & \textbf{97.4} & \textbf{93.0} & \textbf{97.1} & \textbf{89.9} & 95.4 \\
\bottomrule
\end{tabular}}
\vspace{-0.15cm}
\caption{\textbf{Ablation on adaptation.} Except for the versions with the "-GeM" suffix, which utilize GeM features, all other versions use our spatial pyramid representation with the proposed cross-image awareness to yield global features. FrozenDINOv2 and FullTunedDINOv2 represent the use of frozen and fully fine-tuned DINOv2 as backbones, respectively. VanillaAdapter, ConvAdapter, and MulConvAdapter represent the use of a vanilla adapter, 3x3 convolution adapter, and our proposed multi-scale convolution adapter to adapt DINOv2 as the backbone, respectively.}
\vspace{-0.25cm}
\label{tab:ablation_adapt}
\end{table}

\begin{table}
  \centering
  \small
  \setlength{\tabcolsep}{0.4mm}{
  \renewcommand{\arraystretch}{0.92}
  \begin{tabular}{@{}c||ccc||ccc||ccc}
  \toprule
\multirow{2}{*}{Dim}   & \multicolumn{3}{c||}{Pitts30k} & \multicolumn{3}{c||}{Tokyo24/7} & \multicolumn{3}{c}{MSLS-val} \\
\cline{2-10}
& R@1 & R@5 & R@10 & R@1 & R@5 & R@10 & R@1 & R@5 & R@10  \\
\hline
512 & 94.5 & 97.1 & 98.0  & 84.4 & 93.7 & 95.9 & 85.3 & 93.4 &  94.5  \\
1024 & 94.8 & 97.3 & 98.1 & 91.4 & 96.8 & 97.8 & 87.7 & 94.3 &  95.3  \\
2048 & 94.8 & \textbf{97.4} & \textbf{98.2}  & 92.4 & 96.8 & 97.8 & 89.2 & 95.1 & 96.1 \\
4096 & \textbf{94.9} & 97.3 & \textbf{98.2}  & \textbf{93.0} & \textbf{97.5} & \textbf{98.1} & \textbf{90.0} & \textbf{95.4} & \textbf{96.4} \\
10752  & 94.8 & \textbf{97.4} & 98.1 & \textbf{93.0} & 97.1 & 97.8 & 89.9 & \textbf{95.4}  & 96.2 \\
\bottomrule
\end{tabular}}
\vspace{-0.15cm}
\caption{\textbf{Ablation on dimensions of our descriptor.} The original output dimension is 10752.}
\vspace{-0.5cm}
\label{tab:ablation_dim}
\end{table}
\textbf{Impact of descriptor dimensionality.}
In this subsection, we analyze the impact of descriptor dimensionality, and the results are shown in Table \ref{tab:ablation_dim}. Our method gets the best performance when using PCA to reduce the descriptor dimension to 4096-dim, so it is the default dimensionality we recommend. Furthermore, we continue to reduce the dimensionality to observe the point at which performance starts to noticeably decline on each dataset. For Pitts30k, the 512-dim descriptor still achieves an impressive 94.5\% R@1, with no significant decrease compared to the 4096-dim descriptor. However, using the 512-dim descriptor on the other two datasets results in an obvious performance drop. This is mainly due to the drastic condition changes and the perceptual aliasing issue in these datasets, requiring higher-dimensional descriptors to provide sufficient information to distinguish places. When there is a pressing need for low-dimensional descriptors, we suggest using the 1024-dim or 2048-dim descriptor for the place images with obvious condition changes (e.g., Tokyo24/7 and MSLS), the 512-dim descriptor for images like those in Pitts30k.

\textbf{Training time and data efficiency.}
Our model only costs 3.5 hours for training, which is significantly less than the full-day time used by CosPlace/EigenPlaces. The training epochs of ours (10 epochs) are also less than MixVPR (30 epochs) using the same dataset. To further investigate the training time and data efficiency of our method, we reduce the training epoch and training data, and the yielded results are shown in Table \ref{tab:epoch}. When the model is trained with only 10\% of the training data for 1 epoch (i.e., 0.1 epoch), our method achieves better performance than previous methods (except EigenPlaces) on Pitts30k and Tokyo24/7. The training time used is only 0.038h (i.e., 2.3 min). The advantages of our method in data efficiency are mainly due to the fact that the adapter-based method maintains the powerful representation ability of the pre-trained foundation model, while our proposed cross-image encoder is an easy-to-train module.

\begin{table}[!t]
  \centering
  \small
  \setlength{\tabcolsep}{1.3mm}{
  \renewcommand{\arraystretch}{0.92}
  \begin{tabular}{@{}c|c||cc||cc||cc}
  \toprule
\multirow{2}{*}{Epoch}   &  Training  & \multicolumn{2}{c||}{Pitts30k} & \multicolumn{2}{c||}{Tokyo24/7} & \multicolumn{2}{c}{MSLS-val} \\
\cline{3-8}
&time (h)  & R@1 & R@5 & R@1 & R@5 & R@1 & R@5 \\
\hline
10 & 3.5 & \textbf{94.8} & \textbf{97.4} & 93.0 & \textbf{97.1} & \textbf{89.9} & \textbf{95.4} \\
5 & 1.8 & 94.0 & 97.2 & \textbf{93.3} & 96.2 & 89.1 & 95.3  \\
1 & 0.36 & 93.3 & 96.7 & 92.7 & 95.9 & 85.4 & 93.8 \\
\textbf{0.1} & \textbf{0.038} & 92.5 & 96.5 & 88.9 & 96.2 & 79.1 & 88.4 \\
\bottomrule
\end{tabular}}
\vspace{-0.15cm}
\caption{The results of CricaVPR with different training epochs.}
\vspace{-0.5cm}
\label{tab:epoch}
\end{table}

\section{Conclusions}
In this paper, we presented CricaVPR, a robust global representation method with cross-image correlation awareness for VPR. Our method leverages the cross-image encoder equipped with the attention to establish the correlation among multiple images within a batch, enabling the model to harvest useful information from other images while generating the feature representation of an image. This makes the produced global features condition-invariant, viewpoint-invariant, and capable of addressing perceptual aliasing. Furthermore, we proposed a multi-scale convolution-enhanced adaptation method to introduce proper local information and effectively unleash the capability of the pre-trained foundation model for VPR. Experimental results on several VPR benchmark datasets demonstrate that our CricaVPR can provide a robust global representation to address various challenges in VPR and outperforms SOTA methods by a significant margin.

\subsubsection*{Acknowledgments}
This work was supported by the National Key R\&D Program of China (2022YFB4701400/4701402), SSTIC Grant (KJZD20230923115106012), Shenzhen Key Laboratory (ZDSYS20210623092001004), Beijing Key Lab of Networked Multimedia, and the Project of Peng Cheng Laboratory (PCL2023A08).

{
    \small
    \bibliographystyle{ieeenat_fullname}
    \bibliography{main}
}
 
\clearpage
\setcounter{page}{1}
\maketitlesupplementary

\section{Overview}
This supplementary material provides the following additional content about experimental results and analysis:

\vspace{0.18cm}
7. Visualizations of Place Features Using t-SNE

\vspace{0.18cm}
8. Tunable Parameters

\vspace{0.18cm}
9. Additional Results on Challenging Datasets

\vspace{0.18cm}
10. Additional Ablations on Cross-image Encoder

\vspace{0.18cm}
11. Effects of Batch Size

\vspace{0.18cm}
12. Additional Ablations on MulConvAdapter

\vspace{0.18cm}
13. Effects of Adaptation on the Used SPM Feature

\vspace{0.18cm}
14. Comparison to Other Methods with the Same\par\quad\; Training Dataset

\vspace{0.18cm}
15. Datasets Details

\vspace{0.18cm}
16. Compared Methods Details

\vspace{0.18cm}
17. Additional Qualitative Results and Failure Cases

\vspace{0.18cm}
18. Limitations
\vspace{0.18cm}

Note that the experiments in this supplementary material are conducted as in the main paper. That is, PCA is used to reduce the descriptor dimensionality to 4096-dim when comparing our method with other methods. However, it is not used by default in ablation experiments.

\begin{figure*}[!t]
	\centering
	\includegraphics[width=0.9\linewidth]{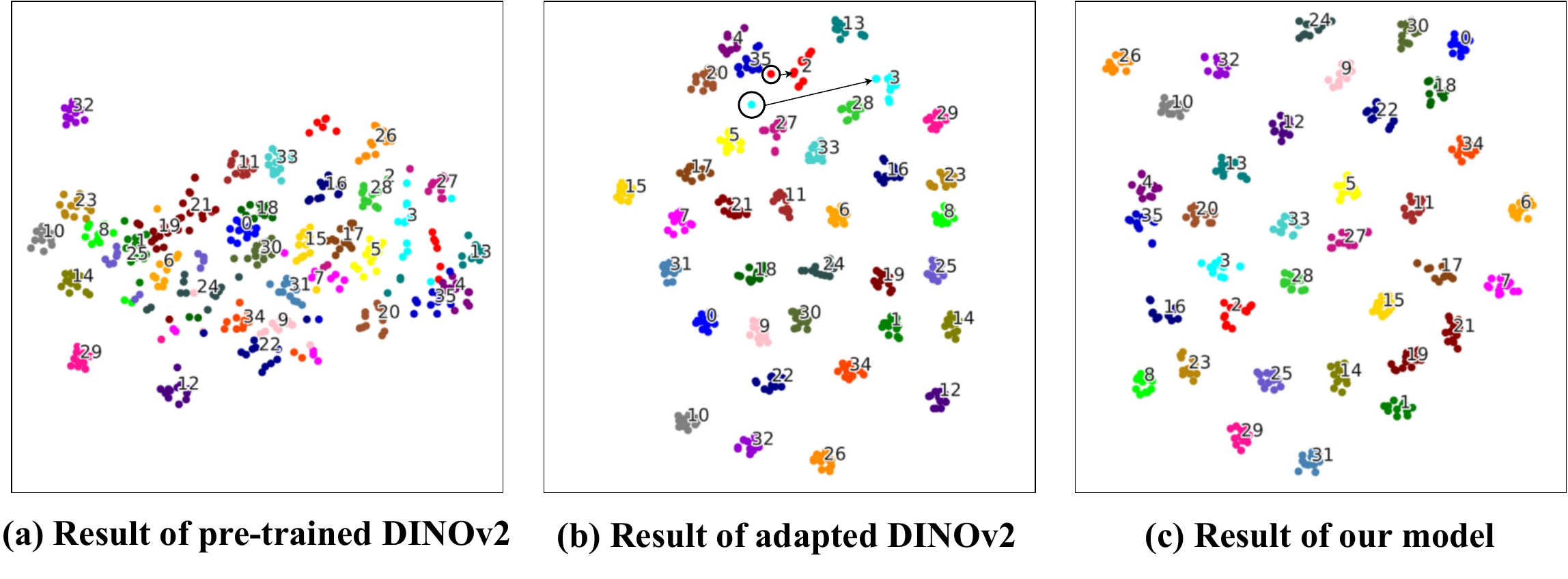}
	\vspace{-0.4cm}
	\caption{
		\textbf{Visualizations of place features in 2-dimensional space using t-SNE.} We use the features of 432 images from 36 different places (i.e. 36 categories) for visualization. (a), (b), and (c) are the results of pre-trained DINOv2, adapted DINOv2 (using our MulConv adapter), and our complete model (using MulConv adapter and cross-image encoder), respectively. Note that the positions of two points in (b) are improper, that is, the corresponding feature representation will suffer perceptual aliasing.
	}
	\label{supfig:distribution}
\end{figure*}

\section{Visualizations of Place Features using t-SNE}
In this section, we use the t-SNE \cite{tsne} method to map our place features to 2-dimensional space and visualize their distribution. We employ pre-trained DINOv2, adapted DINOv2 (with our MulConv adapter), and our entire network (with our MulConv adapter and cross-image encoder) to extract features of 432 images from 36 different places (12 images per place). There exist variations in viewpoints and conditions among the 12 images of the same place. Fig. \ref{supfig:distribution} illustrates the visualization results. It can be observed that some features of different places, which are extracted by pre-trained DINOv2, are not well separated. This demonstrates the limited discriminability of place features extracted by pre-trained DINOv2. However, after performing our adaptation, the adapted DINOv2 successfully distinguishes most of the places, while a few places are still not well distinguished. By applying both our adaptation and the cross-image encoder, our proposed model effectively clusters image features of the same place and separates features of different places, i.e., pulling the features of the same place closer together and pushing the features of different places farther apart. This clearly demonstrates the efficacy of our approach in addressing the challenge of perceptual aliasing.

It is worth mentioning that this visualization method commonly used in classification tasks has rarely been used in previous VPR works. We can use it thanks to the recently proposed GSV-Cities dataset \cite{gsv} (and the SF-XL dataset \cite{cosplace}) that split place images into a finite number of categories.

\section{Tunable Parameters}

We provide detailed model parameters as shown in Table \ref{suptab:param} (using CosPlace as baseline). Since we use an adapter-based parameter-efficient fine-tuning method to train our model, the tunable part of our model only contains the adapter inserted into the backbone and the cross-image encoder after the backbone. The number of tunable parameters of our model is 20.2M, which is only about 1/5 of the full fine-tuning DINOv2 (with the cross-image encoder). This is also less than that of CosPlace using ResNet50 to produce 2048-dim features (including 26.3M tunable parameters in the model and 582.8M tunable parameters of classifiers).
\begin{table}[!t]
  \centering
  \scalebox{0.92}{
  \setlength{\tabcolsep}{1mm}{
  \renewcommand{\arraystretch}{1.1}
  \begin{tabular}{@{}l|c|c|c|c|c}
  \toprule
Method & Total & Backbone & Adapter & Others & \textbf{Tunable}\\
\hline
CosPlace-V & - & 14.7  & 0  & 0.3 & 7.3+145.7  \\
CosPlace-R & - & 23.5  & 0  & 4.2 & 26.3+582.8  \\
\hline
FullTuning & 97.6 & 86.6  & 0  & 11.0 & 97.6  \\
Ours & 106.8 & 86.6+9.2 & 9.2 & 11.0 & 20.2 \\
\bottomrule
\end{tabular}}}
\caption{The number of parameters (M) in models. We mainly focus on tunable parameters (the last column). ``CosPlace-V" and ``CosPlace-R" represent the CosPlace methods using VGG16 and ResNet50 to produce 512-dim and 2048-dim features, respectively. Taking CosPlace-V as an example, since it adds multiple classifiers (for multiple groups of training data) after the model during training, the tunable parameters contain the parameters of the trainable part in the model (7.3M) and all classifiers (145.7M). ``FullTuning" represents full fine-tuning of the DINOv2 backbone (including our cross-image encoder) without the adapter. The ``Others" in the table are the aggregation module for CosPlace, the cross-image encoder for FullTuning and Ours (the parameters of GeM pooling are so few that they can be ignored).}
\label{suptab:param}
\vspace{-0.2cm}
\end{table}

\section{Additional Results on Challenging Datasets}
\begin{table*}
  \centering
  \setlength{\tabcolsep}{0.85mm}{
  \renewcommand{\arraystretch}{1.0}
  \begin{tabular}{@{}l||ccc||ccc||ccc||ccc||ccc}
  \toprule
\multirow{2}{*}{Method} & \multicolumn{3}{c||}{Nordland} & \multicolumn{3}{c||}{AmsterTime} & \multicolumn{3}{c||}{SVOX-Night} & \multicolumn{3}{c||}{SVOX-Rain} & \multicolumn{3}{c}{SVOX-Sun}\\
\cline{2-16}
& R@1 & R@5 & R@10 & R@1 & R@5 & R@10 & R@1 & R@5 & R@10  & R@1 & R@5 & R@10 & R@1 & R@5 & R@10 \\
\hline
SFRS \cite{sfrs} & 16.0&24.1&28.7 & 29.7&48.5&55.6 & 28.6&40.6&46.4 & 69.7&81.5&84.6   & 54.8&68.3&74.1 \\
CosPlace \cite{cosplace} & 58.5 &73.7 &79.4 &38.7&61.3&67.3 &44.8&63.5&70.0 &85.2&91.7&93.8  &67.3&79.2&83.8  \\
MixVPR \cite{mixvpr} & \underline{76.2}&\underline{86.9}&\underline{90.3} & 40.2&59.1&64.6 & \underline{64.4}&\underline{79.2}&\underline{83.1} & \underline{91.5}&\underline{97.2}&\underline{98.1}  & 84.8&93.2&94.7  \\
EigenPlaces \cite{eigenplaces} &71.2&83.8&88.1 &\underline{48.9}&\underline{69.5}&\underline{76.0} &58.9&76.9&82.6 & 90.0&96.4&98.0 & \underline{86.4}&\underline{95.0}&\underline{96.4}  \\
\hline
CricaVPR (ours) & \textbf{90.7}&\textbf{96.3}&\textbf{97.6} & \textbf{64.7}&\textbf{82.8}&\textbf{87.5} & \textbf{85.1}&\textbf{95.0}&\textbf{96.7} & \textbf{95.0}&\textbf{98.2}&\textbf{98.7} &  \textbf{93.7}&\textbf{98.4}&\textbf{98.6} \\
\bottomrule
\end{tabular}}
\caption{Comparison to SOTA methods on challenging datasets. The best is highlighted in \textbf{bold} and the second is \underline{underlined}. We employ PCA to reduce the descriptor dimension of our method to 4096-dim.}
\vspace{-0.2cm}
\label{suptab:compare_SOTA}
\end{table*}

The main paper has presented the R@1 results of our method compared to state-of-the-art (SOTA) methods on three challenging datasets, i.e., Nordland, AmsterTime, and SVOX (SVOX-Night, SVOX-Rain). Here, we provide the complete R@1/R@5/R@10 results as shown in Table \ref{suptab:compare_SOTA}, complementing another challenging query subset (SVOX-Sun) of the SVOX dataset. Before our method was proposed, MixVPR and EigenPlaces had their own advantages on these challenging datasets, and no method completely outperformed the other methods. However, our proposed CricaVPR achieves better performance compared to all previous methods on these datasets, particularly outperforming other methods by a large margin on Nordland, AmsterTime, and SVOX-Night, which are quite difficult.

Moreover, we also provide the results of our CricaVPR on Pitts250k (97.5\% R@1) in Section \ref{sec:trainingdata}.

\section{Additional Ablations on Cross-image Encoder}
\begin{table*}
  \centering
  \scalebox{0.95}{
  \renewcommand{\arraystretch}{1.25}
  \begin{tabular}{@{}l||ccc||ccc||ccc}
  \toprule
\multirow{2}{*}{Cross-image encoder} & \multicolumn{3}{c||}{Pitts30k} & \multicolumn{3}{c||}{Tokyo24/7} & \multicolumn{3}{c}{MSLS-val} \\
\cline{2-10}
& R@1 & R@5 & R@10 & R@1 & R@5 & R@10 & R@1 & R@5 & R@10 \\
\hline
No encoder & 90.6 & 95.9 & 97.2 & 85.1  & 93.3 & 95.6 & 85.5 & 93.2 & 94.3 \\
\hline
Transformer encoder layer $\times$1 & 92.9 & 96.6 & 97.5 & 92.7 & 95.2 & 96.5 & 89.6 & \textbf{95.7} & \textbf{96.4} \\
Transformer encoder layer $\times$2 & \textbf{94.8} & \textbf{97.4} & \textbf{98.1} & \textbf{93.0} & \textbf{97.1} & \textbf{97.8} & \textbf{89.9} & 95.4  & 96.2 \\
Transformer encoder layer $\times$3 & 94.5 & \textbf{97.4} & \textbf{98.1} & \textbf{93.0} & 96.2 & \textbf{97.8} & 88.8 & 94.7 & 96.1  \\
\bottomrule
\end{tabular}}
\vspace{-0.1cm}
\caption{The results of constructing the cross-image encoder using different numbers of transformer encoder layers.}
\label{suptab:encoder}
\end{table*}

\begin{table*}
  \centering
  \scalebox{0.98}{
  \setlength{\tabcolsep}{1.05mm}{
  \renewcommand{\arraystretch}{1.1}
  \begin{tabular}{@{}c|c||ccc||ccc||ccc}
  \toprule
Batch Size & with cross- & \multicolumn{3}{c||}{Pitts30k} & \multicolumn{3}{c||}{Tokyo24/7} & \multicolumn{3}{c}{MSLS-val} \\
\cline{3-11}
(Number of Places) & image encoder & R@1 & R@5 & R@10 & R@1 & R@5 & R@10 & R@1 & R@5 & R@10  \\
\hline
$NP$ = 16 & $\times$ & 89.5 & 95.3 & 96.8 & 75.6 & 89.2 & 91.4 & 80.9 & 90.4 & 92.7  \\
$NP$ = 32 & $\times$ & 89.9 & 95.2 & 96.7  & 81.3 & 91.1 & 93.0 & 83.0 & 93.1 & 93.8 \\
$NP$ = 64 & $\times$ & 90.7 & 95.9 & 97.5  & 84.4 & 94.3 & 96.5 & 84.1 & 92.3 & 94.2 \\
$NP$ = 72  & $\times$ & 90.6 & 95.9 & 97.2 & 85.1  & 93.3 & 95.6 & 85.5 & 93.2 & 94.3 \\
\cline{1-11}
$NP$ = 16 &\checkmark & 94.6 & 97.0 & 97.7 & 87.9 & 94.9 & 96.2 & 84.1 & 92.6 & 94.2  \\
$NP$ = 32 &\checkmark & 94.8 & 97.4 & 98.0  & 91.1 & 94.9 & 96.8 & 85.0 & 93.1 & 95.1 \\
$NP$ = 64 &\checkmark & \textbf{94.9} & \textbf{97.5} & \textbf{98.1}  & 92.4 & 95.6 & 97.1 & 88.1 & 95.0 & 95.1 \\
$NP$ = 72  &\checkmark & 94.8 & 97.4 & \textbf{98.1} & \textbf{93.0} & \textbf{97.1} & \textbf{97.8} & \textbf{89.9} & \textbf{95.4}  & \textbf{96.2} \\
\bottomrule
\end{tabular}}}
\vspace{-0.1cm}
\caption{Results of different training batch sizes, i.e., different numbers of places (4 images per place). $NP$ is the abbreviation of ``Number of Places". We provide the results with or without the cross-image encoder.}
\vspace{-0.2cm}
\label{suptab:batchsize}
\end{table*}

In the main paper, we have combined the proposed cross-image correlation awareness implemented by our cross-image encoder with three different global representations to demonstrate its effectiveness. In this section, we further compare the performance of constructing the cross-image encoder using different numbers of transformer encoder layers, and the results are shown in Table \ref{suptab:encoder}. Compared to not using the cross-image encoder (No encoder), incorporating the cross-image encoder constructed with any number of transformer encoder layers leads to significant performance improvements. However, when only one transformer encoder layer is used, there is still a noticeable performance gap compared to using multiple transformer encoder layers (on Pitts30k and Tokyo24/7), indicating that a single transformer encoder layer alone cannot sufficiently correlate images within a batch. The best performance is achieved when using two transformer encoder layers, which is the recommended configuration.

\section{Effects of Batch Size}
Since our method correlates all images within a batch and utilizes the cross-image variations (including images from the same place and images from different places) as a cue to guide the representation learning in VPR, the training batch size is also a factor that may have an impact on performance. The training dataset GSV-Cities \cite{gsv} provides 4 images per place by default, and we use different batch sizes, i.e., one batch contains different numbers of places, to train our models. It should be noted that we use the multi-similarity (MS) loss to train the models (same as MixVPR), which inherently leads to a result that a larger batch size is more conducive to providing hard sample pairs to train a robust model. Therefore, we also provide the results obtained at different batch sizes without using the cross-image encoder as a reference. The results are shown in Table \ref{suptab:batchsize}. Regardless of whether the cross-image encoder is used, the performance degradation caused by the smaller batch size is not obvious on Pitts30k, but is significant on more difficult Tokyo24/7 and MSLS. When using the cross-image encoder, the absolute R@1 drops caused by using the smallest batch size ($NP=16$) compared to the largest batch size ($NP=72$) on Pitts30k, Tokyo24/7, and MSLS-val are 0.2\%, 5.1\%, and 5.8\% respectively. When the cross-image encoder is not used, the absolute R@1 drops caused by that are 1.1\%, 9.5\%, and 4.6\% respectively. This indicates that: 1) More challenging (test) datasets require larger batch size to train a more robust model. 2) Our proposed cross-image encoder, to some extent, reduces the demand for a larger batch size when using the MS loss for training.

\begin{table}
  \centering
  \scalebox{0.85}{
  \setlength{\tabcolsep}{0.4mm}{
  \renewcommand{\arraystretch}{1.2}
  \begin{tabular}{@{}c||ccc||ccc||ccc}
  \toprule
Batch   & \multicolumn{3}{c||}{Pitts30k} & \multicolumn{3}{c||}{Tokyo24/7} & \multicolumn{3}{c}{MSLS-val} \\
\cline{2-10}
Size & R@1 & R@5 & R@10 & R@1 & R@5 & R@10 & R@1 & R@5 & R@10  \\
\hline
1 & 91.6 & 95.7 & 96.9  & 89.5 & 94.6 & 96.2 & 88.5 & 95.1 & 95.7 \\
4 & 93.9 & 97.2 & 97.7  & 87.3 & 93.7 & 94.6 & 88.0 & \textbf{95.5} & \textbf{96.5} \\
8 & \textbf{94.8} & \textbf{97.4} & \textbf{98.1}  & 91.7 & 96.2 & 97.5 & 89.1 & 95.1 & 95.9 \\
16  & 93.7 & 97.0 & \textbf{98.1} & \textbf{93.0} & \textbf{97.1} & \textbf{97.8} & \textbf{89.9} & 95.4  & 96.2 \\
32  & 93.0 & 96.9 & 97.9 & 92.7 & 96.2 & 97.5 & 88.9 & \textbf{95.5} & 96.2 \\
\bottomrule
\end{tabular}}}
\vspace{-0.1cm}
\caption{Results of different inference batch size.}
\label{suptab:infer_batchsize}
\end{table}
In addition, we also conduct experiments to study the impact of different batch sizes during inference, i.e., inference batch size. The results are as shown in Table \ref{suptab:infer_batchsize}. Since our method learns cross-image correlation-aware representation during training, setting the batch size to 1 during testing makes our cross-image encoder ineffective, further leading to the gap between training and testing, i.e., performance in this case will be reduced. Besides, an inference batch size that is too small (e.g., 4) will lead to unstable results (even worse than when it equals 1). Although the inference batch size that achieves the best performance on different datasets does not appear to be fixed (too small or too large will reduce performance), setting it to 16 can achieve excellent results on all datasets. So we set it to 16 (except on Pitts30k/Pitts250k we set it to 8 for better results).

\section{Additional Ablations on MulConvAdapter}

We have verified the effectiveness of the proposed multi-scale convolution adapter (MulConvAdapter) by comparing it with the vanilla adapter and ConvAdapter (i.e., Convpass \cite{convpass}). To further demonstrate the advantages of MulConvAdapter over adapters using only a single-size convolution kernel, we compare MulConvAdapter with three adapter variants employing three different convolution kernel sizes (1×1, 3×3, and 5×5). To be fair, the three adapters based on a single convolution kernel use skip connection like our MulConvAdapter (the ConvAdapter in the main paper does not), that is, our MulConvAdapter differs from these three adapters only in the convolution kernel. The results are presented in Table \ref{suptab:adapter}. Except for our MulConvAdapter, the adapters based on 1×1, 3×3, and 5×5 convolution kernels have advantages in different datasets (and metrics), indicating that it is difficult for an adapter with a single-size convolution kernel to perform well for all place images on the VPR task. In contrast, our MulConvAdapter integrates these three convolution kernels to consistently provide proper local information, thus achieving the best performance.

\begin{table}
  \centering
  \scalebox{0.85}{
  \setlength{\tabcolsep}{0.4mm}{
  \renewcommand{\arraystretch}{1.2}
  \begin{tabular}{@{}c||ccc||ccc||ccc}
  \toprule
Conv   & \multicolumn{3}{c||}{Pitts30k} & \multicolumn{3}{c||}{Tokyo24/7} & \multicolumn{3}{c}{MSLS-val} \\
\cline{2-10}
Size & R@1 & R@5 & R@10 & R@1 & R@5 & R@10 & R@1 & R@5 & R@10  \\
\hline
1×1 & 94.5 & 97.2 & 97.8  & \underline{91.7} & \underline{95.9} & \underline{97.1} & \underline{88.2} & \underline{95.3} & 95.5 \\
3×3 & 94.3 & 97.1 & \underline{97.9}  & \underline{91.7} & 95.2 & 96.8 & 87.6 & 94.3 & 95.8 \\
5×5 & \underline{94.7} & \underline{97.3} & 97.8  & 90.2 & 94.6 & 96.8 & 87.6 & 95.1 & \textbf{96.4} \\
\hline
MulConv  & \textbf{94.8} & \textbf{97.4} & \textbf{98.1} & \textbf{93.0} & \textbf{97.1} & \textbf{97.8} & \textbf{89.9} & \textbf{95.4}  & \underline{96.2} \\
\bottomrule
\end{tabular}}}
\vspace{-0.1cm}
\caption{The results of convolution-based adapters. ``MulConv" is our MulConvAdapter. Note that the single convolution kernel adapter here has one more skip connection than the 3×3 convolution adapter (ConvAdapter) in the main paper.}
\vspace{-0.1cm}
\label{suptab:adapter}
\end{table}

\section{Effects of Adaptation on the Used SPM Feature}
\begin{table}
  \centering
  \scalebox{0.9}{
  \setlength{\tabcolsep}{1mm}{
  \renewcommand{\arraystretch}{1.2}
  \begin{tabular}{@{}c||cc||cc||cc}
  \toprule
\multirow{2}{*}{Ablated versions}   & \multicolumn{2}{c||}{Pitts30k} & \multicolumn{2}{c||}{Tokyo24/7} & \multicolumn{2}{c}{MSLS-val} \\
\cline{2-7}
& R@1 & R@5 & R@1 & R@5 & R@1 & R@5 \\
\hline
FrozenDINOv2-GeM & 79.2 & 90.1  & 65.4 & 83.8 & 40.8 & 51.5  \\
FrozenDINOv2-SPM & 74.8 & 90.1  & 49.8 & 67.0 & 45.4 & 60.7  \\
\hline
Adapt-GeM & 87.1 & 94.0 & 70.2 & 85.4 & 78.4 & 87.8 \\
Adapt-SPM & 90.6 & 95.9 & 85.1  & 93.3 & 85.5 & 93.2 \\
\bottomrule
\end{tabular}}}
\vspace{-0.1cm}
\caption{The results of the GeM and SPM representation using a frozen DINOv2 or adapted DINOv2 backbone. All results here have been provided in Table 4 and Table 5 of our main paper.}
\vspace{-0.2cm}
\label{suptab:ablation_spm}
\end{table}
In our method, we mainly use the spatial pyramid model (SPM) representation that combines the class token and the GeM feature. An interesting phenomenon is that when using the frozen DINOv2 as the backbone, the SPM feature (FrozenDINOv2-SPM) performs worse than GeM (FrozenDINOv2-GeM) on Pitts30k and Tokyo24/7 (see Table \ref{suptab:ablation_spm}). However, after using our adaptation, Adapt-SPM performs much better than Adapt-GeM. This shows that our adaptation makes the combined class token and GeM features in the SPM representation more compatible. 

\section{Comparison to Other Methods with the Same Training Dataset}
\label{sec:trainingdata}
Most methods (except MixVPR) use different training datasets than our method. The GSV-Cities dataset used in our method has been shown to achieve better results than the datasets with weak supervision (e.g., Pitts30k and MSLS) \cite{gsv}. Training different methods with the same dataset can promote fair comparisons. However, completely achieving it is hard as some methods are designed based on the characteristics of a certain (type of) dataset, and training on others may make some components of them meaningless. To minimize the impact of the training dataset on results, we use the results (reported in the MixVPR paper) of NetVLAD and CosPlace (both based on ResNet50) trained on GSV-Cities for a more fair comparison. The results are shown in Table \ref{tab:gsv} and our method still significantly outperforms others. Note that in this section we have added the results on Pitts250k (larger but easier than Pitts30k). Besides, we also provide the results of training our model on the smallest/weakest Pitts30k dataset in Table \ref{tab:pitts_crica}. Our model trained on Pitts30k still gets SOTA results (better than EigenPlaces trained on SF-XL and much better than SFRS also trained on Pitts30k).

\begin{table}
  \small
  \centering
  \small
  \setlength{\tabcolsep}{0.5mm}{
  \begin{tabular}{@{}c|c||ccc||ccc}
  \toprule
\multirow{2}{*}{Method} & \multirow{2}{*}{Training set} & \multicolumn{3}{c||}{Pitts250k} & \multicolumn{3}{c}{MSLS-val} \\
\cline{3-8}
& & R@1 & R@5 & R@10 & R@1 & R@5 & R@10 \\
\hline
CosPlace\ddag &SF-XL & 92.3 & 97.4 & 98.4  & 87.4 & 94.1 & 94.9 \\
\hline 
NetVLAD\dag &GSV-Cities & 90.5 & 96.2 & 97.4 & 82.6 & 89.6 & 92.0  \\
CosPlace\dag &GSV-Cities & 91.5 & 96.9 & 97.9  & 84.5 & 90.1 & 91.8  \\
CricaVPR &GSV-Cities & \textbf{97.5} & \textbf{99.4} & \textbf{99.7} & \textbf{90.0} & \textbf{95.4} & \textbf{96.4}  \\
\bottomrule
\end{tabular}}
\vspace{-0.1cm}
\caption{The results of methods trained on GSV-Cities. The suffix {\dag/\ddag} means that the method is different from the main paper on the backbone and/or training set. Since SF-XL is built for CosPlace (or it is part of CosPlace), CosPlace{\ddag} trained on SF-XL is better than CosPlace{\dag} trained on GSV-Cities.}
\vspace{-0.1cm}
\label{tab:gsv}
\end{table}

\begin{table}
  \small
  \centering
  \small
  \setlength{\tabcolsep}{0.5mm}{
  \begin{tabular}{@{}c|c||ccc||ccc}
  \toprule
\multirow{2}{*}{Method} & \multirow{2}{*}{Training set} & \multicolumn{3}{c||}{Pitts30k} & \multicolumn{3}{c}{Pitts250k} \\
\cline{3-8}
& & R@1 & R@5 & R@10 & R@1 & R@5 & R@10 \\
\hline
SFRS &Pitts30k & 89.4 & 94.7 & 95.9 & 90.7 & 96.4 & 97.6  \\
\hline
MixVPR &GSV-Cities & 91.5 & 95.5 & 96.3 & 94.1 & 98.2 & 98.9  \\
EigenPlaces &SF-XL & 92.5 & 96.8 & 97.6 & 94.1 & 97.9 & 98.7 \\
CricaVPR* &Pitts30k & \textbf{93.0} & \textbf{96.9} & \textbf{97.9} & \textbf{95.9} & \textbf{99.0} & \textbf{99.5}  \\
\bottomrule
\end{tabular}}
\vspace{-0.1cm}
\caption{Results of CricaVPR* trained on Pitts30k.}
\vspace{-0.2cm}
\label{tab:pitts_crica}
\end{table}

\section{Datasets Details}

\textbf{Pitts30k} \cite{pitts} is derived from Google Street View panoramas with GPS labels. It consists of images from 24 different viewpoints for each place in urban scenes, exhibiting significant viewpoint variations, moderate condition variations, and a small number of dynamic objects. Pitts30k is a subset of Pitts250k (but harder than Pitts250k for most methods). In our experiments, we mainly use the Pitts30k test set.

\textbf{Tokyo24/7} \citep{densevlad} comprises a total of 75,984 database images and 315 query images from urban environments. The query images are selected from a pool of 1,125 images captured from 125 places, each involving 3 different viewpoints and 3 different times of the day. This dataset shows viewpoint variations and significant condition changes, particularly day-night changes.

\textbf{MSLS} (Mapillary Street-Level Sequences) \cite{msls} is a large-scale VPR dataset that encompasses more than 1.6 million images captured in urban, suburban, and natural environments across 30 cities spanning six continents. This dataset provides GPS coordinates and compass angles for each image, and shows various changes caused by illumination, weather, season, viewpoint, dynamic objects, and so on. It is divided into three sets: training, public validation (MSLS-val), and withheld test (MSLS-challenge). To ensure comprehensive evaluation, we assess the model on both the MSLS-val and MSLS-challenge sets, as done in previous works \cite{patchvlad,dhevpr,selavpr}.

\textbf{Nordland} \cite{nordland} captures images from a fixed viewpoint in the front of a train in four seasons. This dataset exhibits significant variations in conditions such as season and lighting, without viewpoint changes. Its images primarily depict suburban and natural environments, and the ground truth information is provided through frame-level correspondence. Following previous works \cite{eigenplaces}, we extract images at 1FPS, and use the winter images as queries and the summer images for reference (i.e. database).

\textbf{AmsterTime} \cite{amstertime} contains more than a thousand query-reference image pairs captured from Amsterdam. Each pair consists of a grayscale historical image as the query and a contemporary image from the same place (identified by human experts) as the reference. The dataset involves very long-term time spans, and diverse domain variations in viewpoints, modalities (RGB vs grayscale), etc., which makes it quite difficult for VPR.

\textbf{SVOX} \cite{SVOX} is a cross-domain VPR dataset collected in a variety of weather and lighting conditions. It includes a large-scale database sourced from Google Street View images spanning the city of Oxford. The queries are extracted from the Oxford RobotCar dataset \cite{robotCar} and divided into multiple subsets for different weather and lighting conditions. We evaluate the model performance using the three most challenging query subsets: SVOX-Night, SVOX-Rain, and SVOX-Sun.

\section{Compared Methods Details}
\textbf{NetVLAD} \citep{netvlad} is a well-known VPR approach with a differentiable VLAD layer, which can be integrated into common neural networks. In our experiments, we use its PyTorch implementation\footnote{\href{https://github.com/Nanne/pytorch-NetVlad}{https://github.com/Nanne/pytorch-NetVlad}} with the released VGG16 model trained on Pitts30k for comparison.

\textbf{SFRS} \cite{sfrs} utilizes self-supervised image-to-region similarities to mine hard positive samples for training a more robust NetVLAD model. In the comparison experiments, we follow its official implementation\footnote{\href{https://github.com/yxgeee/OpenIBL}{https://github.com/yxgeee/OpenIBL}} with the model trained on the Pitts30k dataset. 

\textbf{Patch-NetVLAD} \cite{patchvlad} is a two-stage method that utilizes NetVLAD-based multi-scale patch-level features to re-rank the candidate images retrieved using NetVLAD global features. The official implementation\footnote{\href{https://github.com/QVPR/Patch-NetVLAD}{https://github.com/QVPR/Patch-NetVLAD}} with the performance-focused configuration is used in our experiments. Following the original paper, the model trained on the Pitts30k dataset is tested on Pitts30k and Tokyo24/7, while the model trained on the MSLS dataset is evaluated on MSLS (-val and -challenge).

\textbf{TransVPR} \cite{transvpr} is a two-stage VPR method that leverages attentions from three levels of Transformer to produce global features for candidates retrieval, and employs an attention mask to filter feature maps to yield key-patch descriptors for re-ranking candidates. The official implementation\footnote{\href{https://github.com/RuotongWANG/TransVPR-model-implementation}{https://github.com/RuotongWANG/TransVPR-model-implementation}} is used for comparison experiments. The model trained on the Pitts30k dataset is evaluated on Pitts30k and Tokyo24/7, and the model trained on the MSLS dataset is assessed on MSLS.

\textbf{CosPlace} \cite{cosplace} treats VPR model training as a classification problem and trains the model on the individually constructed San Francisco eXtra Large (SF-XL) datasets with the Large Margin Cosine Loss (i.e., cosFace) to achieve remarkable results. We follow its official implementation\footnote{\href{https://github.com/gmberton/CosPlace}{https://github.com/gmberton/CosPlace}} with the VGG16 backbone (producing 512-dim global features) for testing.

\textbf{GCL} \cite{gcl} uses an automatic annotation strategy producing graded similarity labels for image pairs to re-label VPR datasets, and a novel generalized contrastive loss to utilize such labels to train contrastive networks. we use the results (yield by the version using ResNet152-GeM with PCA) from the original paper for comparison.

\textbf{MixVPR} \citep{mixvpr} introduces a novel holistic feature aggregation approach for global-retrieval-based VPR. It utilizes feature maps yielded by a pre-trained backbone as initial feature representations, and employs a sequence of Feature-Mixer modules to incorporate global relationships into each feature map to produce final global features. We follow the official implementation\footnote{\href{https://github.com/amaralibey/MixVPR}{https://github.com/amaralibey/MixVPR}} and its best configuration, i.e., using the ResNet50 backbone producing 4096-dim global features, for comparison experiments.

\textbf{EigenPlaces} \cite{eigenplaces} can be seen as an improvement work on CosPlace. This work trains the networks on images from different viewpoints (of the same place), thus improving the viewpoint robustness of learned global representations. It is the most recent work and achieves the best performance on most VPR datasets. We follow its official implementation\footnote{\href{https://github.com/gmberton/EigenPlaces}{https://github.com/gmberton/EigenPlaces}} and the configuration using ResNet50 as the backbone to yield 2048-dim features.

Besides, the results on the three challenging datasets (Nordland, AmsterTime, and SVOX) are directly referenced from the EigenPlaces paper \cite{eigenplaces}. These results are basically consistent with what we have reproduced.

\section{Additional Qualitative Results and Failure Cases}
\begin{figure*}[!t]
	\centering
	\includegraphics[width=0.92\linewidth]{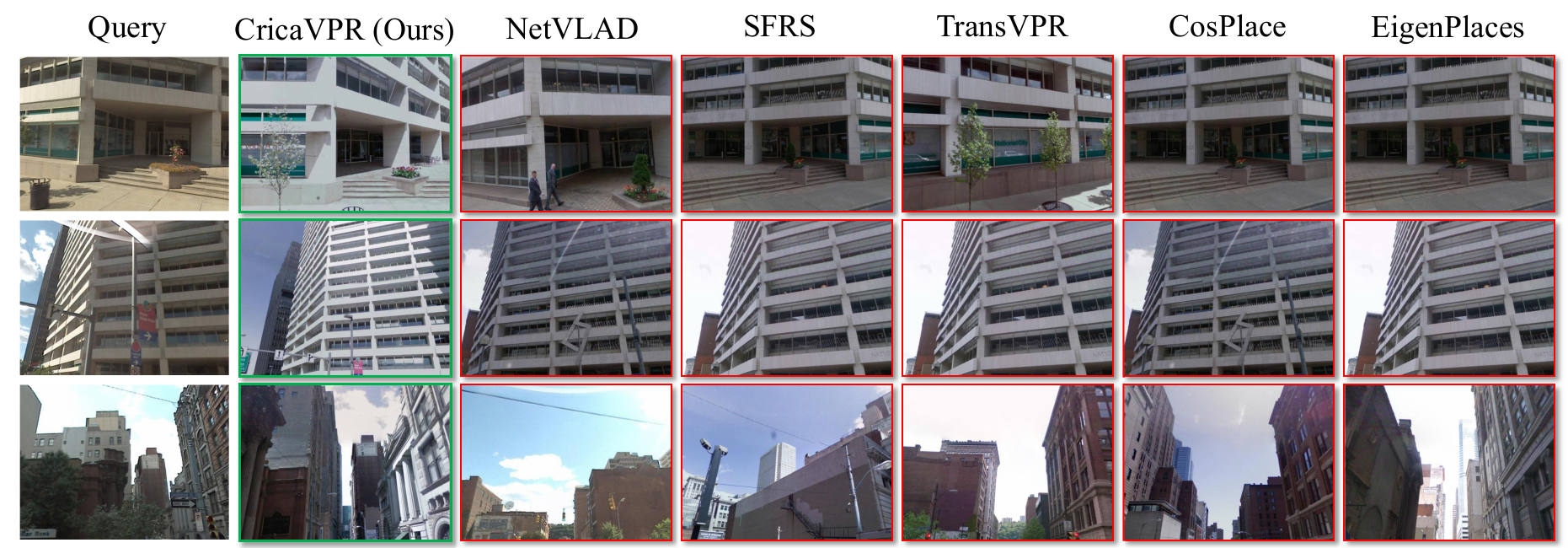}
	\vspace{-0.3cm}
	\caption{
		\textbf{Qualitative results on Pitts30k.} The proposed CricaVPR returns the correct database images, while other methods produce wrong results. In these examples, most of the other methods suffer from perceptual aliasing. In the first two examples, all other methods return highly similar but wrong places. In the third example, the buildings on the right of the images returned by TransVPR, CosPlace, and EigenPlaces are highly similar to the building on the right of the query image, indicating that the appearance of this building is not distinguishable enough, making these methods suffer from perceptual aliasing.
	}
	\vspace{-0.4cm}
	\label{supfig:pitts30k}
\end{figure*}

\begin{figure*}[!t]
	\centering
	\includegraphics[width=0.92\linewidth]{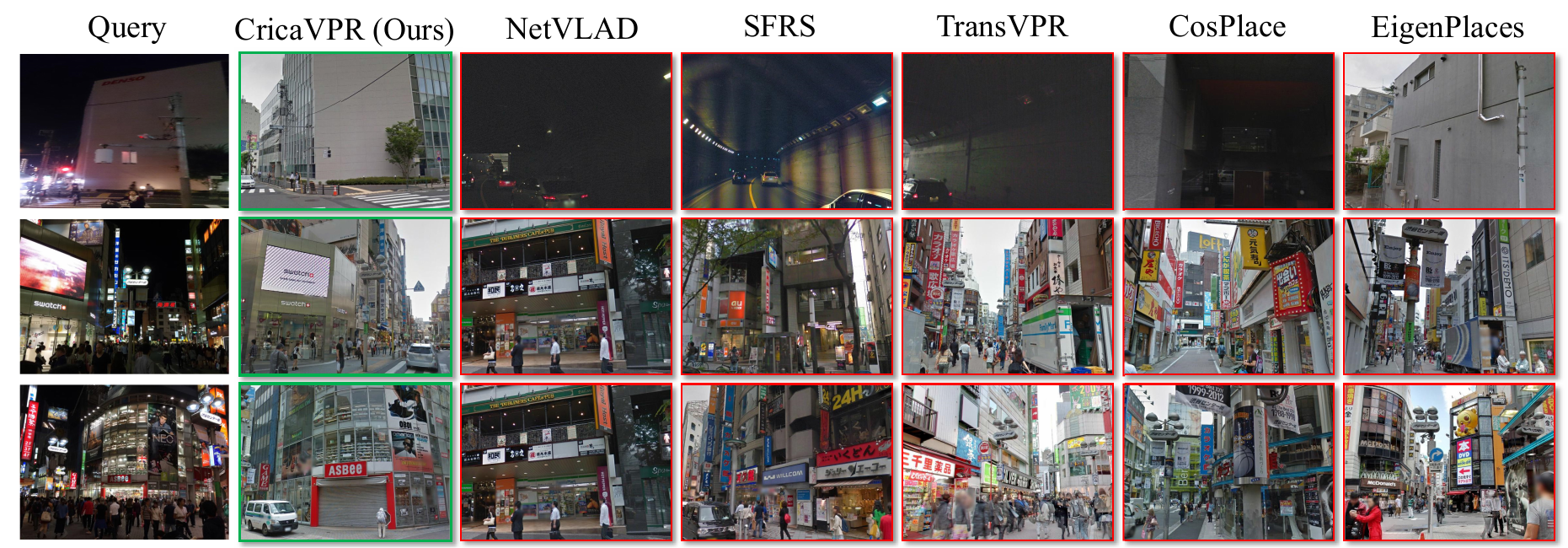}
	\vspace{-0.3cm}
	\caption{
		\textbf{Qualitative results on Tokyo24/7.} The proposed CricaVPR returns the correct database images, while other methods produce wrong results. In these examples, the main challenges are the variations in lighting conditions across day and night, as well as perceptual aliasing. In the first example, as the query image is a nighttime image, all methods except for ours and EigenPlaces return nighttime but incorrect images. EigenPlaces returns a similar but incorrect image.
	}
	\vspace{-0.4cm}
	\label{supfig:tokyo247}
\end{figure*}

\begin{figure*}[!t]
	\centering
	\includegraphics[width=0.92\linewidth]{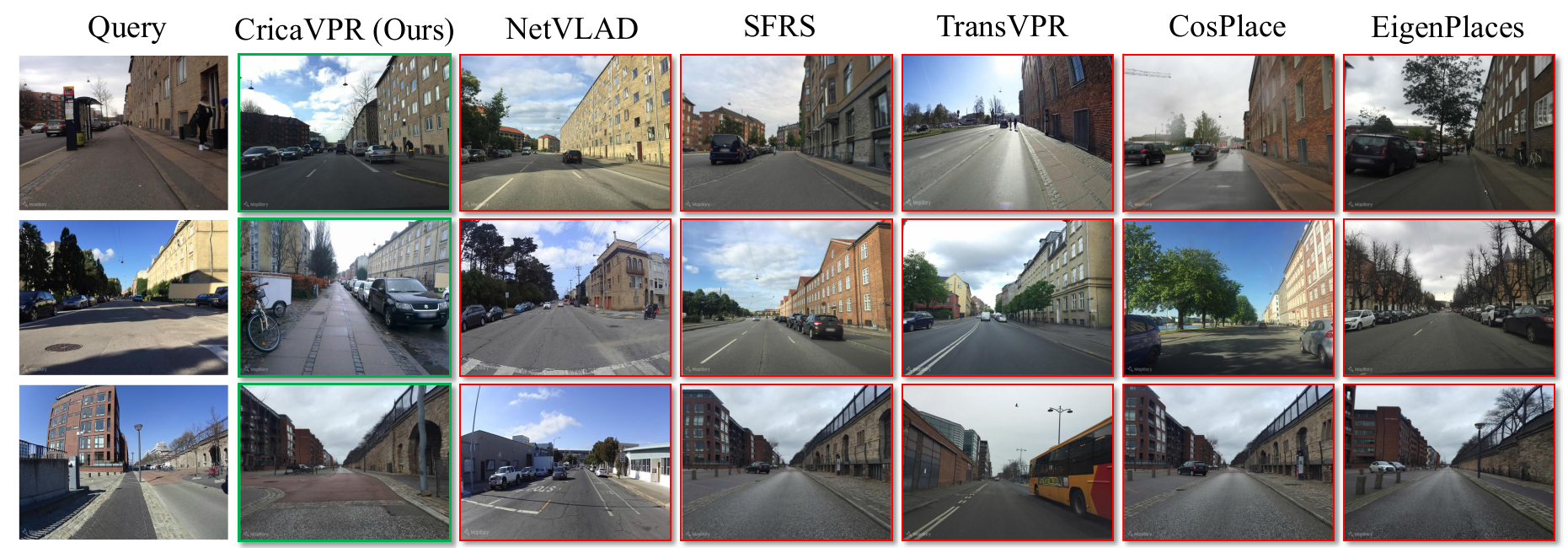}
	\vspace{-0.3cm}
	\caption{
		\textbf{Qualitative results on MSLS-val.} The proposed CricaVPR returns the correct database images, while other methods produce wrong results. In the first example, NetVLAD returns a highly similar but wrong image. In the second example, the building on the left of the query image is occluded by trees, causing NetVLAD, TransVPR, and CosPlace to return incorrect results with obvious trees on the left side. In the third example, SFRS, CosPlace, and EigenPlaces return database images that are geographically close to the query image but exceed the set threshold (i.e. still wrong).
	}
	\vspace{-0.4cm}
	\label{supfig:msls}
\end{figure*}

\begin{figure*}[!t]
	\centering
	\includegraphics[width=0.92\linewidth]{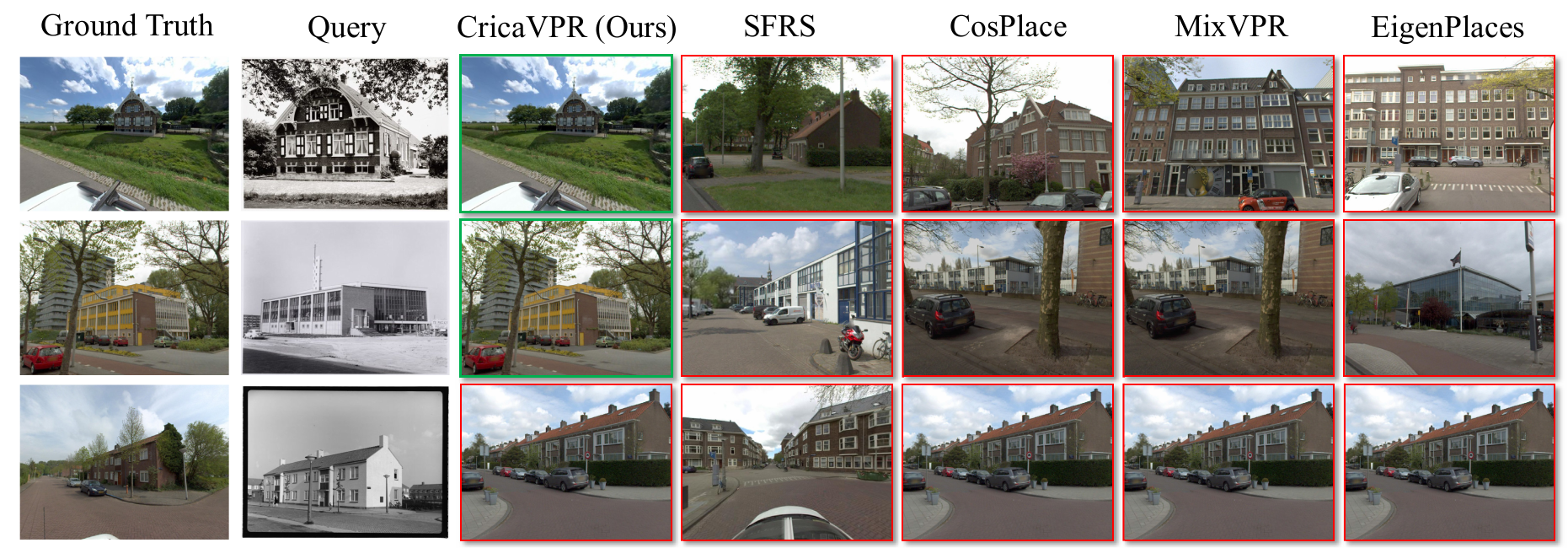}
	\vspace{-0.3cm}
	\caption{
		\textbf{Qualitative results on AmsterTime.} There is a very long time span between the query (grayscale) image and the reference (RGB) image in this dataset. In the first two examples, the proposed CricaVPR returns the right database images, while other methods produce wrong results. In the first example, the discriminative buildings only occupy a small region of the reference image. In the second example, a new building appears in the reference image, and the original building has undergone some modifications. These cause other methods to return incorrect results. In the last example, there are images from different places in the database that are highly similar to the query image, causing none of the methods to retrieve the correct result.
	}
	\vspace{-0.4cm}
	\label{supfig:amstertime}
\end{figure*}

\begin{figure*}[!t]
	\centering
	\includegraphics[width=0.92\linewidth]{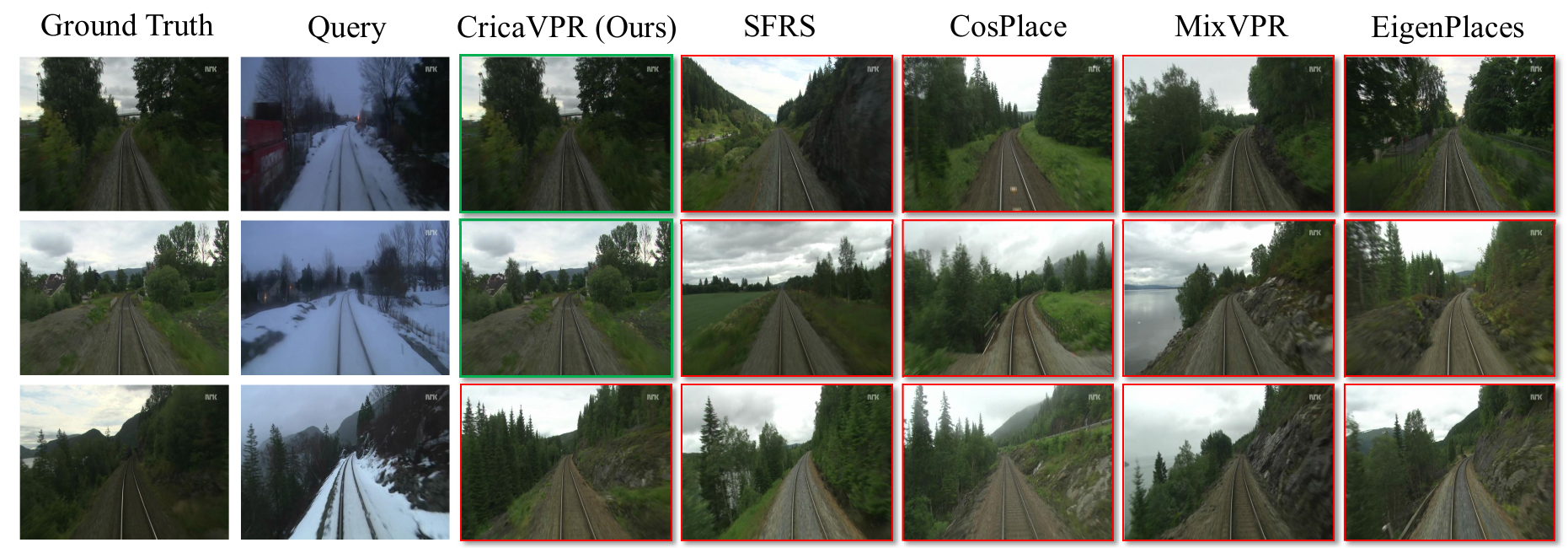}
	\vspace{-0.3cm}
	\caption{
		\textbf{Qualitative results on Nordland.} These examples show drastic variations in conditions (season, weather, and lighting). Meanwhile, there are almost no discriminative buildings in the images. These challenges are difficult to address for previous VPR methods, resulting in incorrect results being returned by all of them. Our method gets the right result in the first two examples but fails in the last one.
	}
	\vspace{-0.4cm}
	\label{supfig:nordland}
\end{figure*}

In the main paper, we have presented a small number of qualitative results to show the robustness of our approach in challenging scenarios. In this section, we add more examples to vividly demonstrate the performance of VPR methods. Fig. \ref{supfig:pitts30k}, Fig. \ref{supfig:tokyo247}, and Fig. \ref{supfig:msls} show examples on Pitts30k, Tokyo24/7, and MSLS-val, respectively. These examples demonstrate that our method is more robust against variations in conditions and viewpoints, as well as perceptual aliasing, than previous methods. Fig. \ref{supfig:amstertime} and Fig. \ref{supfig:nordland} show examples on AmsterTime and Nordland, which demonstrate that our method can correctly recognize place images over long time spans and under extreme environments in general. However, our method also produces erroneous results in a few cases (the last examples of these two figures) when images from different places are very similar, especially when lacking discriminative landmarks. 

\section{Limitations}
In addition to the failure case mentioned in the previous section, our approach has two limitations. First, although our approach achieves excellent results with the 512-dim compact feature on Pitts30k, it does not perform well on datasets with severe condition changes (e.g., Tokyo247, MSLS) when the descriptor dimension is reduced to very low. Secondly, setting the inference batch size to 1 will render our cross-image encoder ineffective, resulting in a gap between training and testing, and thus not achieving optimal performance. These are the focal points for future improvement of our method.

\end{document}